\pdfoutput=1

\documentclass[11pt]{article}

\usepackage{ACL2023}

\usepackage{times}
\usepackage{latexsym}
\usepackage{graphicx}
\usepackage{enumerate}
\usepackage{enumitem}
\usepackage{booktabs}
\usepackage{colortbl}
\usepackage{multirow}
\usepackage{pdfpages}

\usepackage{graphicx} %
\usepackage{booktabs} %
\usepackage{tabularx} %
\usepackage{multirow} %
\definecolor{lightgreen}{RGB}{223,255,219}
\definecolor{lightred}{RGB}{255,219,219}
\definecolor{lightblue}{RGB}{219,235,255}
\definecolor{lightorange}{RGB}{255,235,219}
  
\newcommand{\up}{\cellcolor{lightblue}}  
\newcommand{\dn}{\cellcolor{lightorange}} 

\definecolor{lightgreen}{RGB}{223,255,219}
\definecolor{lightred}{RGB}{255,219,219}
\definecolor{redish}{RGB}{255,102,102}
\definecolor{blueish}{RGB}{31, 78, 192}
\definecolor{orangeish}{RGB}{255, 178, 102}
\definecolor{mypink1}{rgb}{0.858, 0.188, 0.478}
\definecolor{mypink2}{RGB}{219, 48, 122}
\definecolor{mypink3}{cmyk}{0, 0.7808, 0.4429, 0.1412}
\definecolor{mygray}{RGB}{220, 220, 220}
\definecolor{darkbluee}{RGB}{0,17, 113}
\definecolor{purpleNew}{RGB}{151, 45, 204}
\definecolor{purplebg}{RGB}{229, 199, 244}
\definecolor{violet}{rgb}{0.70,0.05,0.65}
\newcommand{\rfive}{\cellcolor{green}}
\newcommand{\rfour}{\cellcolor{lightgreen}}
\newcommand{\rthree}{\cellcolor{yellow}}
\newcommand{\rtwo}{\cellcolor{orangeish}}
\newcommand{\rone}{\cellcolor{redish}}

\definecolor{orangeish}{HTML}{FF9933}
\definecolor{maroon}{HTML}{FF00FF}
\definecolor{greenish}{HTML}{82B366}
\makeatother

\usepackage{comment}
\newcommand\scalemath[2]{\scalebox{#1}{\mbox{\ensuremath{\displaystyle #2}}}}
\newcommand{\myparagraph}[1]{\noindent\textbf{#1}}
\usepackage{amsmath}

\usepackage[T1]{fontenc}

\usepackage[utf8]{inputenc}

\usepackage{microtype}

\usepackage{inconsolata}
\usepackage{comment}
\newsavebox\myv

\usepackage{array}
\makeatletter
\newcolumntype{V}[1]{>{\topsep=0pt\@minipagetrue}p{#1}<{\vspace{-\baselineskip}}}
\makeatother
\newcolumntype{L}[1]{>{\raggedright\let\newline\\\arraybackslash\hspace{0pt}}m{#1}}

\usepackage{listings}
\lstdefinestyle{base}{
  emptylines=1,
  breaklines=true,
  linewidth=\textwidth,
  commentstyle=\color{gray},
  basicstyle=\scriptsize\ttfamily,
  moredelim=**[is][\color{red}]{@}{@},
}
\usepackage{longtable}
\usepackage[ruled,vlined]{algorithm2e}
\SetArgSty{textup}
\newcommand{\dataset}{RDTE}
\newcommand{\datasetlong}{RDTE (Recognizing Decompositional Textual Entailment)}
\newcommand{\sysname}{\textsc{TreeWise}}
\newcommand{\sysnamefull}{\textsc{TreeWise}: \textbf{T}extual \textbf{R}easoning \textbf{E}ngine with \textbf{E}nriched \textbf{W}ays to \textbf{I}ntelligently \textbf{S}earch for \textbf{E}ntailment}
\newcommand{\nellie}{\textsc{Nellie}}

\title{Enhancing Systematic Decompositional Natural Language Inference \\ Using Informal Logic}
\author{Nathaniel Weir$^\spadesuit$\thanks{\ \ Work done in part as intern for Allen Institute for AI.}  \quad Kate Sanders$^\spadesuit$ \quad Orion Weller$^{\spadesuit}$ \quad Shreya Sharma$^\spadesuit$ \\ 
\textbf{Dongwei Jiang}$^{\spadesuit}$ \quad 
\textbf{Zhengping Jiang}$^{\spadesuit}$ \quad
\textbf{Bhavana Dalvi Mishra}$^{\dag}$ \quad \textbf{Oyvind Tafjord}$^\dag$ \quad \\ \textbf{Peter Jansen}$^{\ddag\dag}$  \quad 
\textbf{Peter Clark$^{\dag}$}  \quad \textbf{Benjamin Van Durme$^{\spadesuit}$} \\
$^\spadesuit$Johns Hopkins University
\quad $^\dag$Allen Institute for AI \quad $^\ddag$University of Arizona \\
\texttt{\{nweir, ksande25, vandurme\}@jhu.edu} 
}

\begin{document}
  
\maketitle

\begin{abstract}
Recent language models enable new opportunities for structured reasoning with text, such as the construction of intuitive, proof-like textual entailment trees without relying on brittle formal logic~\cite{tafjord-etal-2022-entailer,weir-etal-2023-nellie}. 
However, progress in this direction has been hampered by a long-standing lack of a clear protocol for determining what \emph{valid compositional entailment} is.
This absence causes noisy datasets and limited performance gains by modern neuro-symbolic engines. 
To address these problems, we formulate a {consistent} and {theoretically grounded} approach to annotating decompositional entailment and evaluate its impact on LLM-based textual inference. 
We find that our new dataset, \datasetlong{}, has a substantially higher internal consistency (+9\%) than prior decompositional entailment datasets.
We also find that training an RDTE-oriented entailment classifier via knowledge distillation and employing it in an entailment tree reasoning engine significantly improves both accuracy and proof quality, illustrating the practical benefit of this advance for textual inference.
\end{abstract}
\begin{figure}[t!]
    \centering
    \includegraphics[width=.95\columnwidth,trim={1mm 2mm 1mm 4mm},clip]{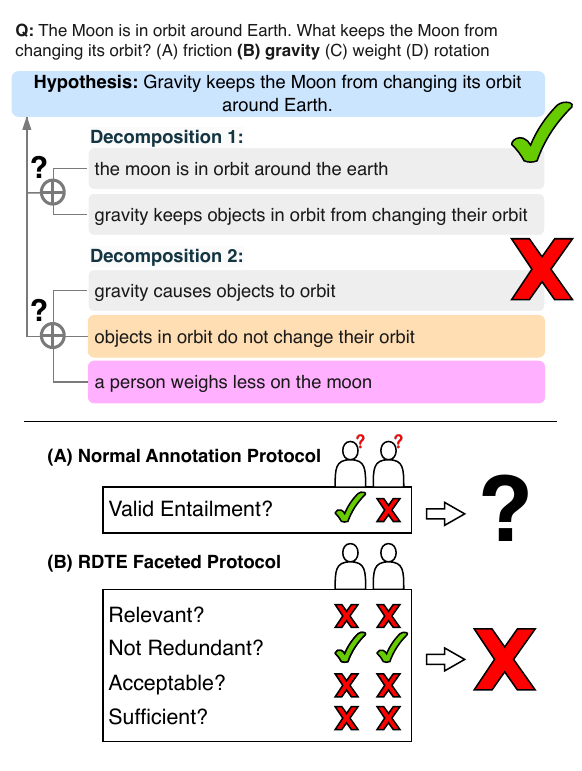}
    \caption{\textbf{(Upper)} Two hypothesis decompositions suggested by an LLM. The first
    makes an argument that is generally acceptable to a human. The second contains \textcolor{orangeish}{a fact that is not always true} and another that is \textcolor{maroon}{irrelevant to the entailment}. Recognizing such an invalid decomposition is core to recent neuro-symbolic reasoning algorithms, but LLMs struggle at the task. \textbf{(Lower)} Ambiguous definitions of entailment have hampered progress in annotating data to improve the models. We find that a faceted definition yields both a clean dataset (RDTE) and significant downstream task improvements.
    }
    \label{fig:cartoon_decomp}
\end{figure}

\section{Introduction}
Textual inference using entailment trees has emerged as a promising avenue for explainable AI in domains requiring complex reasoning~\cite{dalvi-etal-2021-explaining, bostrom-etal-2022-natural}. 
Inspired by classical symbolic reasoners that produce proof trees of a hypothesis,
recent entailment tree systems (e.g. \nellie{}~\cite{weir-etal-2023-nellie} or IRGR~\cite{ribeiro-etal-2022-entailment}) explain a hypothesis by constructing,
via an LM-based generative search algorithm, 
a tree of recursive natural language (NL) decompositions whose children entail their parents.
Entailment trees provide a natural and granular medium for explaining complex multi-step decisions to users
and create opportunities to ground reasoning to large NL knowledge corpora~\cite{weir-etal-2023-nellie}.

Entailment tree reasoners are most useful to humans if they are \textit{right for the right reasons}: They not only give correct answers, but each tree step is a valid deduction supporting its hypothesis.
While some tree generators leverage classifiers for recognizing textual entailment~\cite[RTE;][]{dagan-etal-2005-pascal} to identify which hypothesis decompositions are illogical and should be discarded~\cite{yang-etal-2022-generating}, it is still common for systems to arrive at the correct conclusion while presenting a flawed decomposition, or to accept incorrect conclusions based on faulty logic.
Thus, the task of \textit{recognizing decompositional textual entailment} is a major barrier to progress on tree-based reasoners.

Work on entailment trees has largely avoided this critical question of {reasoning validity}. Existing evaluations focus on tree reconstruction~\cite{ribeiro-etal-2023-street} and QA performance, avoiding considerations of whether the quality of model-generated trees aligns with the accuracy of the answers.
This is largely because reasoning quality is difficult to assess. Evidence suggests humans think of deductive validity less stringently than formal symbolic axioms; they accept explanations 
that are more or less incomplete~\cite{sulik-etal-2021-explanations,tan-2022-diversity}.
Attempts to annotate datasets~\cite{tafjord-etal-2022-entailer, clark-etal-2023-barda} consequently suffer from inconsistent labels, as 
opinions differ between annotators over what constitutes validity.
Without careful instruction, it is hard for annotators to determine where to set their threshold for acceptability on the spectrum from ``strict deductive inferences'' to ``fallible common-sense inferences''~\cite{gubelmann-etal-2023-capturing}.
Thus, asking for a single binary decompositional entailment judgment yields a difficult and noisy annotation task (\autoref{fig:cartoon_decomp}, (A)). 
This leads us to pursue a more consistent protocol for measuring the goodness of one decomposition over another.

Existing discourse surrounding NLI (e.g. \citet{manning-2006-local}) has not touched on the sort of decompositional entailments encountered in a recursive tree search algorithm. Towards addressing this gap,
we propose an evaluation centered on the notion that \textbf{a valid decomposition is a valid argument for why the hypothesis should be believed}. 
We take inspiration from the ``Relevance, Acceptability, and Sufficiency'' criteria for a logically good argument developed in the field of informal logic~\cite{johnson-blair-1977-logical, groarke-2022-informal}, and 
design an approach to assessing entailment in a principled manner. 
We annotate items on ordinal scales targeting different facets of argument analysis, yielding a well-defined and consistent process.
We use this protocol to collect over 1000 expert annotations.
Through experiments, we find that models trained on previous compositional entailment datasets and LLMs like GPT fall short of human performance on the dataset, which we term \datasetlong{}.

We also find that our annotation protocol can serve as a powerful LM prompting mechanism:
Given the same instructions as our human annotators, GPT-4~\cite{openai-2023-gpt4} can serve as a ``teacher'' in a knowledge distillation pipeline that annotates reasoning traces extracted from a tree search algorithm. 
We release a large artifact (24K items per domain) of GPT-4's annotations for this purpose.
We illustrate the effectiveness of student models trained via this pipeline as the linchpin of \sysname{}, a novel entailment tree engine that performs hypothesis proving via decompositional entailment search.
Inspired by the recent \nellie{} engine~\cite{weir-etal-2023-nellie}, \sysname{} leverages in-context learning, forward chaining, branch consolidation, and most importantly, improved {decompositional entailment recognition}. 
It surpasses tree-producing approaches on established benchmarks like EntailmentBankQA, but also adapts to complex tasks such as HotpotQA that require reasoning over less structured knowledge sources like Wikipedia. Using the knowledge-distilled models as reasoning verifiers improves \sysname{}'s QA accuracy \textit{and} raises the quality of the trees it produces. %
Our contributions are thus:
\begin{itemize}[leftmargin=*]
\setlength\itemsep{0em}
\vspace{-2mm}
\item \textbf{A new entailment challenge set}, \dataset{}, crafted via an informal logic-inspired protocol, tasking models to validate hypothesis decompositions
\item \textbf{A knowledge distillation pipeline} that teaches student models to discriminate \textbf{what} and \textbf{why} decompositions are invalid in a given domain.
\item \textbf{A new entailment tree-generating inference engine}, \sysname{}, which outperforms existing tree-based QA methods while generating higher-quality trees in the process. This engine shows to benefit from RDTE knowledge distillation.\footnote{Data, code, and models can be found at 
\url{https://github.com/JHU-CLSP/treewise}.
}
\end{itemize}
\section{Decompositional RTE}
\subsection{Task Definition}
An entailment tree comprises one or multiple hypothesis decompositions.
Each constitutes one instance of decompositional RTE, which we define as follows:
Given the question $Q$, hypothesis $h$, and premises $[p_1, \dots, p_n]$, would proving the premises be contextually sufficient to prove $h$?

As described below, we collect different ordinal scores for each item using criteria from informal logic. 
Each facet informs the ultimate sufficiency label, for which we evaluate models in \S\ref{sec:evaluation}.

\subsection{Background: RAS Criteria}
A seminal framework developed by informal logicians to replace strict deductive logic criteria is known as RAS: \textbf{Relevance}, \textbf{Acceptability}, and \textbf{Sufficiency}~\cite{johnson-blair-1977-logical}. 
Much like RTE work, 
RAS assesses whether a hypothesis can be accepted on the basis of a list of premises.
Noting that each element is subject to substantial academic debate (different systems of informal logic define them differently), we review each criterion as we interpret them for this work:

\myparagraph{Relevance} of premise A concerning conclusion B is defined as whether the truth of A makes a difference to the truth of B.~\cite{blair-2012-relevance}. The extent of this difference can vary; e.g. whether (A) ``the earth has oxygen'' is true technically has relevance to (B) birds can fly (since birds breathe oxygen), but is less relevant than (A') birds have wings.
Relevance is of particular interest to a recursive reasoner, as one would not want to waste time proving an irrelevant statement like (A) when proving (B).

\myparagraph{Acceptability} of a premise is normatively ``worthy of acceptance,'' which can mean either its ostensible truth value or-–in the absence of universal factuality--that in the relevant context, the arguer and the argument recipient accept it to be true. This introduces a hiccup for recursive reasoning algorithms, for which the factuality of a decomposition's premises is commonly determined via search \textit{after} validating the decomposition itself. We ultimately choose to annotate one subset of \dataset{} for premise factuality while not doing so for the other.

\myparagraph{Sufficiency} is ``the property of an argument’s premises of supplying all the grounds that are needed to make it reasonable to believe its conclusion.''~\cite{johnson-blair-1977-logical}. This is left intentionally vague; \citet{blair-2012-relevance} notes ``the criterion of sufficiency, for justificatory arguments, is best seen as a placeholder for whatever version and standards of sufficiency are appropriate for the particular situation in question.'' In this way, we should consider sufficiency to be a question and problem-specific criterion:
\textbf{the grounds for valid entailment are inherently domain-specific}.
\citet{blair-2012-relevance} notes the dependent relationship between sufficiency and the other two criteria: sufficient {presumes} acceptability and relevance.   

\subsection{Implementing RAS for RTE Annotation}
We draw inspiration from these criteria to construct a protocol that investigates a decomposition like the ones shown in \autoref{fig:cartoon_decomp} for each RAS component in turn. 
An important aspect of these normative terms is that they are all scalar: a premise can be more or less relevant and acceptable, and the sufficiency of an argument composed of premises could always be strengthened by adding more premises. 
In a departure from works such as \citet{tafjord-etal-2022-entailer} who collect binary factuality and ``reasoning correctness'' judgments, we collect RAS judgments on an ordinal scale~\cite{zhang-etal-2017-ordinal} from 1 to 5, where each score is assigned a specific set of conditions.
We also collect an ordinal judgment for \textbf{redundancy}, which is not strictly a component of RAS assessment; we observe that redundant premises are particularly problematic for entailment tree search, as proving the same information twice wastes search budget. We implement redundancy as ``conditional irrelevance'': 
\begin{itemize}[leftmargin=*]
\setlength\itemsep{0em}
\vspace{-2mm}
\item If removing a premise \textit{in isolation} doesn't change the extent of the entailment, then it's \textbf{irrelevant}.
\item If removing a premise \textit{in the presence of the other decomposition premises} does not change the extent of the entailment, then it's \textbf{redundant}.\footnote{An irrelevant premise is thus by definition also redundant.}
\end{itemize} \vspace{-1mm}
The sufficiency label most directly resembles that of a typical RTE label, and is what we evaluate models on in later sections.
Following \citet{blair-2012-relevance}'s observation, the sufficiency judgment
is higly dependent on the other criteria;
our annotation directions, found in \S\ref{app:ras_directions}, include a substantial list of 30+ conditions around a decomposition that indicate what the sufficiency score should be based on other criteria scores, e.g. 
``Did you give any fact a 3/5 or lower for redundancy, factuality, or relevance? (Yes = max 4 sufficiency)''.

\section{Data Collection}

We seek to construct a dataset of decompositional entailment judgments that are of the kind that a reasoning system might come across while performing QA. 
Following \citet{tafjord-etal-2022-entailer}, we found it apt to annotate decompositions generated during a model's reasoning search traces over training questions. We use the backward chaining process of the tree search algorithm \sysname{}, which is introduced in \S\ref{sec:nellie_gpt}. 
Each item is a hypothesis and a set of 2-3 premises that the model proposes might conjunctively entail the hypothesis in the context of a given question.
To collect a representative sample of hypotheses, we pull from three classes:
\begin{enumerate}[leftmargin=*]
\setlength\itemsep{0em}
    \item \textbf{Top-level correct hypotheses} representing the right answer options for multiple-choice questions. These are important to annotate so that models are \textit{right for the right reasons.}
    \item \textbf{Recursive correct hypotheses} LLM-generated to decompose the top-level correct hypotheses 
    \item \textbf{Top-level incorrect hypotheses} representing the incorrect answer deemed closest to correct by GPT-4. These are important to annotate so that models are \textit{not wrong for the wrong reasons.}
\end{enumerate}

We collect a mixture of GPT-4 and ChatGPT-generated decompositions using a suite of different styles of prompts described in \S\ref{sec:prompts}. 
We generated decompositions for hypotheses\footnote{We created hypotheses by declarativizing~\cite{demszky-etal-2018-transforming} answer options using GPT-4. For HotpotQA, we first had GPT-4 synthesize incorrect answer options.} in two task domains: multiple choice science QA in \textbf{ARC}~\cite{clark-etal-2018-think} and multi-hop QA over Wikipedia in \textbf{HotpotQA}. While Hotpot questions are constructed over public world knowledge, the esoteric individual facts (e.g. the birth year of some actor) are not common knowledge among humans, and it would be a stretch for the audience of an argument made by a Hotpot decomposition to be expected to know each premise's factuality. 
 We thus {do not annotate Hotpot decompositions for factuality}, but annotate the other facets as normal. 

\subsection{Annotation Process }
Four of this paper's authors, all highly proficient or native English speakers, annotated 1000 decompositions total with two-way redundancy. One author annotated all items. We first annotated several examples together to develop a list of 30+ condition checks that indicate certain RAS scores. We then annotated the rest of the dataset independently.

While the dataset is annotated for sufficiency on a 5-point scale, we found a threshold of $\geq 4$ fitting to evaluate models under binary entailment metrics. 
This corresponds to items for which the only acceptable flaws are (A) a 3rd redundant premise in an otherwise perfect 2-premise entailment or (B) minor missing information that does not affect lay reasoning.
To arrive at a single clean entailment label for evaluation, we reconciled disagreements by discussing all items for which its two annotated sufficiency scores were on either side of 3.5. 
A vast majority disagreements were due to human error, e.g. missing a condition or not recognizing a particular flaw in reasoning. 

\myparagraph{Comparison to Existing Data }
Prior to reconciling disagreements, we measured raw annotation agreement and compared it to recent attempts to collect decompositional entailment labels. \citet{tafjord-etal-2022-entailer} collected annotations for 3.7K decompositions generated by \textbf{Entailer}, while \citet{clark-etal-2023-barda} did so for 24.4K items generated by \textbf{GPT3}.\footnote{The BaRDa dataset comprises a specifically filtered subset of these two datasets that exhibit maximal annotator agreement. We obtained the pre-filtered data from the authors for the purposes of comparison with our work.} 
The instructions given to annotators for these datasets are very high-level, simply asking whether the ``reasoning goes wrong.'' We believe this creates a much lower threshold for acceptability than the \dataset{} protocol; these datasets are majority labeled ``entailment,'' but on manual inspection we found numerous items that exhibited redundant, irrelevant, and fallacious arguments. 

Nevertheless, \textbf{RDTE has a higher internal annotator consistency rate} than these datasets, with a raw agreement rate 9\% higher and Krippendorf $\alpha$ 19 points higher than the previous best.
\vspace{2mm}

{
    \centering
    \scalebox{0.85}{
\begin{tabular}{lccc}
\toprule 
& \textbf{RDTE (ours)} & \textbf{Entailer} & \textbf{GPT3}  \\ \midrule
  Raw agreement   &  \textbf{.79} & .70 & .61 \\
 Krippendorf $\alpha$    & \textbf{.53} & .34 & .31 \\ \bottomrule
\end{tabular}
}
}

\begin{figure}[t!]
    \centering
    \includegraphics[width=\columnwidth,trim={2mm 2mm 1mm 1mm},clip]{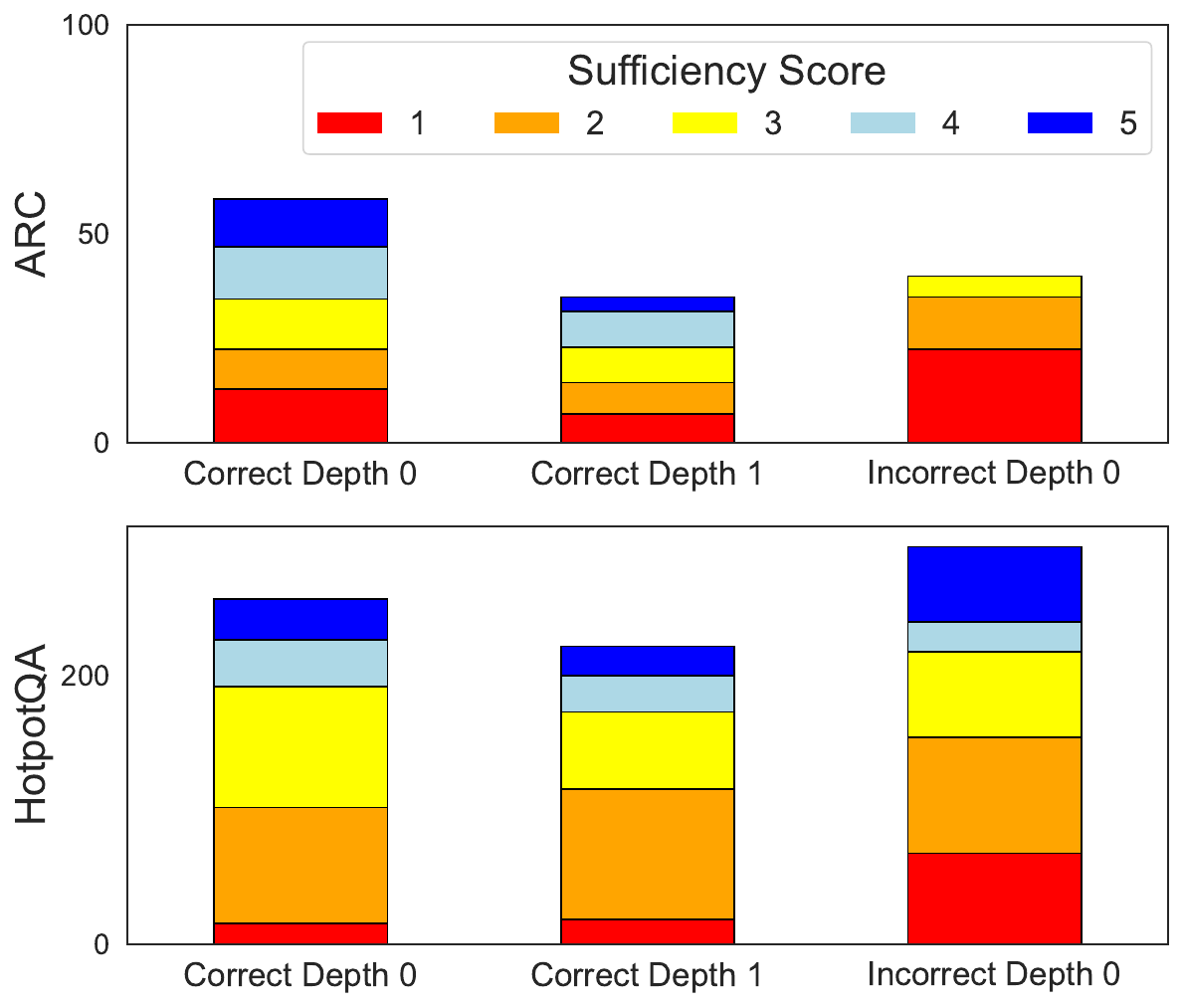}
    \caption{Distribution of the 1000 entailment labels in \dataset{}. Instead of binary entail/non-entailment, we annotate on a 5-point ordinal scale. To evaluate binary judgment models, we treat $\geq$4 as positively labeled.}
    \label{fig:rdte_distribution}
\end{figure}

\begin{figure}[t!]
    \centering
    \scriptsize
    \begin{tabular}{p{5cm}rrr}
    \toprule
    & \textbf{Fa} & \textbf{Rel} & \textbf{Red} \\ \midrule
    \multicolumn{4}{p{7cm}}{\textbf{Q:} The 2005 film Remedy featured Frank Q: Vincent from The Sopranos and several mob movies by which acclaimed director? (A) Jonathan Demme, (B) Martin Scorsese, (C) Ralph De Vito, (D) Steven Hilliard Stern} \\
    \textbf{H}: The 2005 film Remedy featured Frank Vincent from The Sopranos and several mob movies by the acclaimed director Ralph De Vito.\\
    \quad \textbf{P1}: Frank Vincent appeared in The 2005 film Remedy.  & -- & 5 & 5 \\
    \quad  \textbf{P2}: Frank Vincent appeared in The Sopranos and several mob movies & -- & 5 & 5 \\
    \quad  \textbf{P3}: Ralph De Vito directed several mob movies. & -- & 5 & 5 \\
    \multicolumn{4}{p{7cm}}{\textbf{Sufficiency: 3}. The facts are relevant and nonredundant, but they do not link the mob movies by De Vito to those that featured Vincent. Substantial missing information is a 3.}  \\ \midrule 
    \multicolumn{4}{p{7cm}}{\textbf{Q:} Which process best explains how the Grand Canyon became so wide? (A) folding, (B) erosion, (C) deposition, (D) sedimentation} \\
    \textbf{H}: Erosion best explains how the Grand Canyon became so wide.\\
    \quad \textbf{P1}: Changes in the Grand Canyon's landscape include becoming wider.  & 5 & 5 & 5 \\
    \quad  \textbf{P2}: Erosion is one of the processes that can change a landscape. & 5 & 5 & 5 \\
    \quad  \textbf{P3}: The Grand Canyon is a landscape. & 5 & 3 & 5 \\
    \multicolumn{4}{p{7cm}}{\textbf{Sufficiency: 2}. The facts are relevant and nonredundant, but they do not establish a direct causal relationship between erosion and the Grand Canyon becoming wider. Removing P3 does not strongly impact the extent of the entailment. Redundant P + missing information is a 2. }  
    \\ \midrule
    \multicolumn{4}{p{7cm}}{\textbf{Q:} Is Ordos City more west than Yangzhong? (A) No, (B) Same longitude, (C) yes} \\
    \textbf{H}: Ordos City is located in the western part of China.\\
    \quad \textbf{P1}: Ordos City is predominantly rural.  & - & 3 & 5 \\
    \quad  \textbf{P2}: Predominantly rural areas in China are often found in the western part of the country. & - & 3 & 5 \\
    \multicolumn{4}{p{7cm}}{\textbf{Sufficiency: 2}. The argument commits a fallacy of division by assuming that because predominantly rural areas in China are often in the west, Ordos City, being rural, must also be in the western part. This reasoning does not adequately support the conclusion without specific geographic evidence. Fallacious reasoning and irrelevant premises is a 2.}
    \\ \bottomrule
    \end{tabular}
    \caption{Example \dataset{} annotations.
    }
    \label{tab:my_label}
\end{figure}

\subsection{RDTE Analysis}

\autoref{fig:rdte_distribution} depicts the breakdown of sufficiency labels within the \dataset{} ARC (267) and Hotpot (775) decompositions. While the ordinal scores are relatively well distributed, only 27\% of the dataset is labeled a 4 or 5, creating a large binary label imbalance. This statistic highlights the importance of performing well on this task: 3 out of every 4 decompositions generated by ChatGPT and GPT-4 did not pass our sufficiency rubric.

49\% of RDTE items had at least one premise rated as irrelevant ($\leq 3/5$); 36\% had at least one rated as redundant. 28\% of the items with a factuality rating had at least one nonfactual premise.
We note that \textit{none of the incorrect ARC hypothesis decompositions were labeled as sufficient}; this highlights that the RDTE protocol, when including the factuality facet, does a comprehensive job of systematically ruling out decompositions for which there is necessarily some issue (else the hypothesis must be correct). Contrastly, there \textit{are} positively-labeled decompositions of incorrect hypothesis for Hotpot, because we did not annotate for factuality. There are thus decompositions that, were their premises true, would entail a wrong hypothesis.

\section{RDTE Evaluation}
\label{sec:evaluation}
We elicit sufficiency judgments over the 1000 gold-annotated RDTE items from a series of methods based on the RDTE protocol as well as existing methods for judging decompositional entailment. 
We test them on the gold sufficiency scores converted to binary labels by assiging 4 and 5s to positive entailment.
Factuality, relevance, and redundancy predictions were not directly evaluated.

\myparagraph{GPT Methods }
We test the following prompt-based methods, using both GPT-4 and ChatGPT.\footnote{\texttt{gpt-4-0613} and \texttt{gpt-3.5-turbo-1106}} All prompts can be found in the appendix.
\begin{itemize}[leftmargin=*]
\setlength\itemsep{0em}
\vspace{-2mm}
    \item An \textbf{ICL} prompt containing the RDTE annotation rubric and 4 example batches (10-15 decompositions per hypothesis) and a \textbf{Zero-Shot} prompt containing only the RDTE rubric.
    \item The prompts used by \citet{clark-etal-2023-barda} to evaluate models on the \textbf{BaRDa} dataset. These are separate prompts for the entailment and premise factuality judgments (see \autoref{fig:barda-prompt}).
\end{itemize}
\myparagraph{Existing Methods }
We test various RTE models from recent  entailment tree-based reasoners: the T5-based \textbf{Entailer-11B}~\cite{tafjord-etal-2022-entailer}
 and the similar \textbf{\nellie{}-3B} model~\cite{weir-etal-2023-nellie}, as well as the fine-tuned RoBERTa classifier used by the \nellie{} system as an entailment filter. 
We also take an off-the-shelf (\textbf{OTS}) RoBERTa finetuned for NLI using a suite of 600 datasets\footnote{huggingface.co/sileod/deberta-v3-large-tasksource-nli} that includes QASC~\cite{khot-etal-2020-qasc} and ARC.

\myparagraph{Knowledge Distillation on Silver Data }
Search algorithms like \sysname{} make hundreds of calls to NLI models for every question, making large, slow models like GPT-4 unrealistic for use as an entailment filter. We instead explore using knowledge distillation to train smaller student models to imitate GPT-4's performance. 
We extracted 24K RDTE judgments from GPT-4 over 2.3K hypotheses in each of the ARC and Hotpot domains. 
We release these large silver datasets for future work.\footnote{GPT-4 obtains a .61 and .44 Spearman $\rho$ with humans on the ARC and HotpotQA RDTE splits, respectively.}

We used the silver data to fine-tune (A) the RoBERTa model used in the \nellie{} engine, and (B) ChatGPT using OpenAI's fine-tuning API. We trained the former to predict the binarized silver sufficiency judgments and the latter to replicate GPT-4's exact answers to the RDTE prompt.
For token efficiency, the GPT-4 teacher and ChatGPT student perform classification over a batch of decompositions for a single hypothesis in each prompt.\footnote{Batching barely affects performance on RDTE (\autoref{tab:rdte_full_appendix}).}

\begin{table}[t!]
\small
\centering
\scalebox{.9}{
\setlength{\tabcolsep}{5pt}
\begin{tabular}{lll}
\toprule
 & \multicolumn{1}{l}{\textbf{ARC}} & \multicolumn{1}{l}{\textbf{Hotpot}} \\ 
\midrule
\multicolumn{1}{l}{\textbf{Prompted Methods}}  \\
GPT-4 (RDTE ICL) & \textbf{59} & \textbf{53} \\
GPT-4 (RDTE Zero-Shot)  & 58 & 49 \\
GPT-4 (BaRDa) & 44 & 48 \\
ChatGPT (RDTE ICL) & 36 & 35 \\
ChatGPT (RDTE Zero-Shot) & 40 & 38$^\dagger$ \\
ChatGPT (BaRDa)  & 43$^\dagger$ & 34 \\  \midrule
{\textbf{T5 and Cross Encoders}}  \\
Entailer-11B & 48 & 38 \\
NELLIE-3B  & 43 & 36 \\
NELLIE RoBERTa Filter & 37 & 36 \\
OTS NLI RoBERTa & 45$^\ddagger$ & 44$^\ddagger$  \\ \midrule 
\multicolumn{1}{l}{\textbf{Knowledge Distillation}} \\  
ChatGPT & 48 (+5)$^\dagger$ & 51(+13)$^\dagger$  \\
RoBERTa & \textbf{66} (+21)$^\ddagger$ & \textbf{56}(+12)$^\ddagger$ \\
\bottomrule
\end{tabular}}
\caption{RDTE entailment results (F$_{0.5}$ score). We find that RDTE prompting of GPT-4 greatly outperforms existing approaches to compositional entailment, including BaRDa prompting and fine-tuned approaches. Knowledge disillation from GPT-4 (RDTE Zero-Shot) improves student models by 5-21 points,
at times outperforming GPT-4 itself.
 $\dagger/\ddagger$ denote the student models and the best performing analogous non-distilled models.}

\label{tab:rdte_results_main}
\end{table}

\subsection{RDTE Results}
Our metric is \textbf{F-score ($\beta=0.5$)}, putting double precedence on precision over recall. Precision is particularly crucial for our needs:
false positives in a backward-chaining search can quickly create error propagation, wasted time on invalid search branches, and worse trees. Contrastly, while recall is important, it is less catastrophic to overfilter valid decompositions than to underfilter bad ones. 

\autoref{tab:rdte_results_main} shows F$_{0.5}$ performance by models on the RDTE dataset. We also display precision and recall in \autoref{tab:rdte_full_appendix}. We observe that \textbf{no model cracks 70}, suggesting room for future improvement. Overall we find the \textbf{RDTE-oriented prompts to GPT-4 outperform all existing methods for compositional entailment} by 10 or more points.\footnote{\autoref{tab:rdte_results_main} shows GPT-4's score when thresholding on \textbf{5}, not \textbf{4}, which we found to score lower (see \autoref{tab:rdte_full_appendix}). }

We find that knowledge distillation proves effective on \dataset{}: the student ChatGPT models improve 5-13\% over the closest ChatGPT methods, while the \textbf{fine-tuned RoBERTa student ultimately outperforms the teacher GPT-4 method}. \autoref{tab:rdte_full_appendix} shows that this is due to higher precision (68 to 55) at the expense of recall (57 to 90). 
The RoBERTa student is the {most} precise classifier tested, and the ChatGPT student is the next-highest among non-GPT-4 models; however, both show much lower recall than the others tested.

\begin{figure}
    \centering
    \includegraphics[width=\columnwidth]{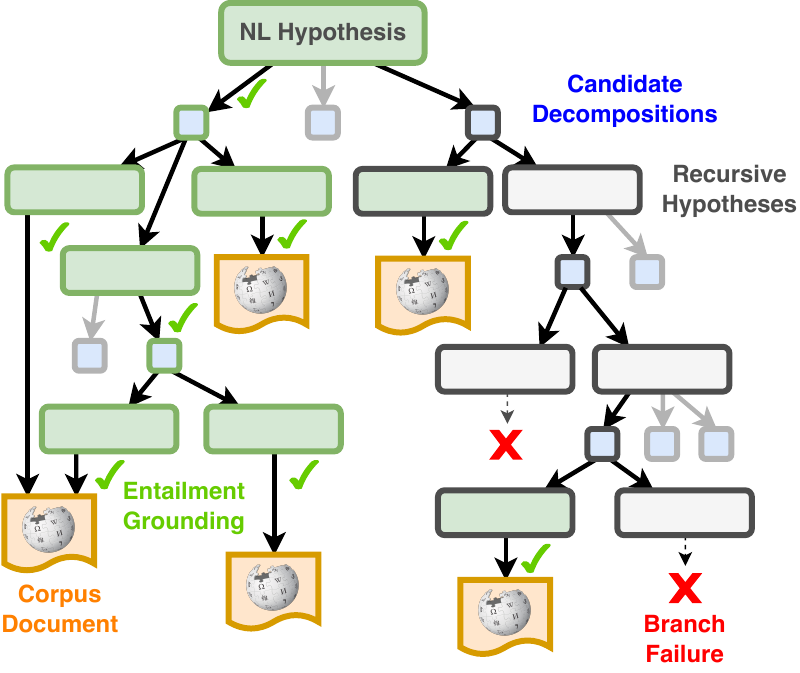}
    \caption{%
    \sysname{} generates many \textcolor{blue}{premise decompositions} of a hypothesis and checks whether any candidates are valid entailments. Premises are then recursively decomposed until it finds any \textcolor{greenish}{tree(s)} fully grounded in \textcolor{orange}{one or more documents from a corpus like Wikipedia}. Statements entailed by documents are generated via \textit{forward chaining}, while the rest of the search is \textit{backward}.
    \textcolor{gray}{Many decompositions end up untraversed} due to the search budget or nonentailment.
    }
    \label{fig:search_trace}
\end{figure}
\section{\sysname{}}
\label{sec:nellie_gpt}

To illustrate the impact that an RDTE-based entailment model can have on systematic reasoning, we apply it as a module in a new, state-of-the-art entailment engine called \sysnamefull{}. \sysname{} builds upon the backward chaining entailment tree search framework first introduced by \nellie{}~\cite{weir-etal-2023-nellie}. 
The core functionality of \sysname{} is to answer the question, \textbf{``is NL hypothesis $H$ compositionally entailed by a corpus of documents $C$ in the context of a question $Q$?''}
Illustrated in \autoref{fig:search_trace}, the search attempts to ground the hypothesis via recursive decomposition into premises entailed by passages in a verified corpus such as Wikipedia.
Implemented in Prolog, the algorithm follows a breadth-first strategy, performing the following at each recursive hypothesis $H$:
\begin{enumerate}[leftmargin=*]
	\setlength\itemsep{0em}
 \vspace{-1mm}
	\item \textbf{Retrieve} a set of $s$ support documents from a corpus like Wikipedia that are likely to contain information relevant to the hypothesis. Check if any of these documents entails $H$ using a series of single-premise entailment classifiers. If so, $H$ is proved. See \S\ref{app:retrieval} for implementation details.%
	\item \textbf{Check whether $H$ is a paraphrase} of a previously considered hypothesis $H'$ using SBERT~\cite{reimers-gurevych-2019-sentence}. If so, 
 we associate it with the 
 proof branches of $H'$. %
	\item \textbf{Forward generate} $i$ inferences from the support documents via an instruction-tuned LLM.
	\item \textbf{Decompose} $H$ into a set of $d$ candidate decompositions using a series of prompts to the LLM. We use a diverse set of prompts to propose candidates that encourage using the $i$ inferences. %
	\item \textbf{Condense the decompositions} by deduplicating semantically equivalent candidates.
	\item \textbf{Filter the decompositions} using a series of decompositional entailment classifiers to weed out those that would not logically support $H$. %
 
	\item \textbf{Recur on the premises} to continue the search.
\end{enumerate}
This process is repeated until either a user-specified expansion budget is exhausted, or a proof is found and no search branches remain that could yield a higher score.
The system is designed to be amenable to a costly API; it batches together decompositions of a single hypothesis into one prompt, then asks it to score them all at once.

\sysname{} represents an overhaul of the T5-based \nellie{} system, incorporating a variety of prompting methods, forward chaining, and the capacity to reason over long passages.  
We discuss the search logic and architectural comparison in \S\ref{app:nellie_vs_treewise}.

\section{\sysname{} Experiments}
\begin{table*}[ht!]
\small
    \centering
    \scalebox{1}{
    \begin{tabular}{rc|cccccc}
\toprule
& & \multicolumn{2}{c}{EBQA on WorldTree} & \multicolumn{2}{c}{EBQA on Wikipedia} & \multicolumn{2}{c}{HotpotQA on Wikipedia} \\ \cmidrule(lr){3-4} \cmidrule(lr){5-6} \cmidrule(lr){7-8}
 \textbf{Method}    & \textbf{W/ Silver?} & \textbf{QA} & \textbf{Tree Integrity} & \textbf{QA} & \textbf{Tree Integrity} & \textbf{QA} & \textbf{Tree Integrity}  \\ \midrule 
 \sysname{} & \up Yes & \textbf{79.2} & \textbf{75.2} & \textbf{73.2} & \textbf{74.7 }& \textbf{51.3} & \textbf{66.6} \\
  & \dn No & 72.8 & {71.6} & 69.0 & 71.9 & 46.1 & 62.9 \\
   Stepwise Generator & \up Yes & 66.4 &  69.5 & 56.8 & 68.3 & 47.5 & 58.8 \\
 & \dn No & 63.7 & 68.1 & 51.5 & 64.7 & 49.4 & 58.6\\
 End-to-End Generator & \up Yes & 68.7 & 66.4 & 57.6 & 66.4 & 48.5 & 59.1 \\
 & \dn No & 68.7 & 66.1 & 55.1 & 64.0 & 43.0 & 57.7 \\

\bottomrule
 \end{tabular}
 }
    \caption{QA and tree integrity score for tree-generating approaches with vs without silver knowledge distillation.}
    \label{tab:qa_results}
\end{table*}
The distillation framework suggests a way forward for improving entailment engines in new domains: first extract non-optimized reasoning traces from the engine, annotate them under the RDTE protocol with GPT-4 (which is itself too slow and expensive to use in the engine), train student models on the silver data, and then substitute the student as an entailment classifier in the engine. 
The experiments below emulate this scenario and show that the resulting engines improve on both QA and on generated tree quality, highlighting the effectiveness of applying RDTE-trained distillation student models to entailment tree creation algorithms. 

\myparagraph{Entailment Tree QA Task Definition }
We pursue a version of evidence-grounded explainable QA in which models are not only right, but right for the right reasons.
Given corpus $C$, multiple-choice question $Q$
 and hypotheses $h_1, \dots, h_m$
 corresponding to answer options $a_1, \dots, a_m$, we task models to predict correct answer $a_i^{*}$ 
 and generate entailment tree $T_{h_i}$ rooted at $h_i$
 such that (1) the $T_{h_i}$’s leaves are in $C \cup Q$
 and (2) the tree's steps explain how each parent follows compositionally from its children.
Models are evaluated on selecting $a_i^{*}$ as well as the correctness of the reasoning in $T_{h_i}$.

\subsection{QA Evaluation}

\myparagraph{Datasets }
We evaluate \sysname{} and a series of entailment tree-creating baselines on the two QA datasets used to construct RDTE: 340 test questions from the EntailmentBank (\textbf{EBQA}), and 419 HotpotQA questions recast as multiple choice by using GPT-4 to generate incorrect answer options.

As \sysname{} can flexibly hook up to different types of knowledge sources, we evaluate it on EBQA using two different scenarios:
(1) EBQA using the clean factbase WorldTree~\cite{xie-etal-2020-worldtree} as the knowledge source, and (2) EBQA using an index over English Wikipedia as the knowledge source.
For HotpotQA, we only use that task's specific Wikipedia index as the knowledge source.

\myparagraph{Metrics }
We take a two-pronged approach to evaluating systems: they should produce both strong QA accuracy \textbf{and} coherent and logically sound entailment trees explaining the chosen answer.
To evaluate the latter, we introduce a new model-based \textbf{tree integrity score}. We use GPT-4 to score each of a tree's component entailment steps under the RDTE protocol. 
Following the intuition that an argument is only as strong as its weakest link, we take the minimum score as the tree's overall score. 

\myparagraph{Baselines }
We compare a version of \sysname{} using the RDTE-trained student ChatGPT and RoBERTa models to an identical version that uses an ICL prompted ChatGPT and the RoBERTa model used by \nellie{}, which is trained on non-RDTE entailment data. We use ChatGPT as the decomposition generator and QA2D model for \sysname{}, meaning it does not use GPT-4 for anything at test time. We also evaluate a pair of greedy baselines for entailment tree creation. These baselines are designed to mimic \sysname{}'s behavior in a simplified manner without the systematic search:

\begin{itemize}[leftmargin=*]
\setlength\itemsep{0em}
\vspace{-2mm}
\item An \textbf{end-to-end} tree generator that retrieves one set of facts and then uses ChatGPT to decode an entailment tree in one fell swoop.
\item A \textbf{stepwise} tree generator that iteratively retrieves support facts and then decodes one step of the entailment tree until the tree is fully grounded or a maximum number of steps (10) is reached.  
\end{itemize}

Psuedocode for these is provided in \S\ref{app:baselines}. Each process is repeated 5 times, yielding 5 different candidate trees for each answer choice, then fed to the \textbf{tree integrity} scorer using the student ChatGPT to score each tree. We take the highest-scoring tree and corresponding answer as the final output. We compare these to non-RDTE versions where the student ChatGPT is replaced by regular ChatGPT.

\myparagraph{Results }
\autoref{tab:qa_results} shows QA results for these methods. We observe that for all methods, tree integrity score increases when using the knowledge distilled student. For all but 1 baseline on HotpotQA, QA accuracy also goes up by 1 to 7 points. 
We observe that \sysname{} achieves the highest QA and integrity scores, while the stepwise outperforms the end-to-end generator under integrity but vice versa for two QA scenarios.
See \autoref{fig:more-trees-arc} and \ref{fig:more-trees-hotpot} for example trees generated by \sysname{} using Wikipedia as its knowledge source. 
\section{Related Work}

\myparagraph{Annotating Textual Entailment Datasets}
The PASCAL RTE Challenge~\cite{dagan-etal-2005-pascal}, SNLI~\cite{bowman-etal-2015-large} and MNLI~\cite{williams-etal-2018-broad} have led to numerous NLI leaderboards. %
These datasets (A) are generally not tied to downstream reasoning tasks; and (B) commonly target {uncontroversial} items for which annotators have high agreement without needing granular instructions. 
The BaRDa dataset~\cite{clark-etal-2023-barda} throws out a large fraction of decomposition annotations because of low agreement. In contrast, \dataset{} specifically targets the hard-to-evaluate items that unavoidably manifest during systematic reasoning algorithms. 
Our use of ordinal annotation draws from JOCI~\cite{zhang-etal-2017-ordinal}, a collection of common sense inferences annotated on a 5-point scale of likeliness.

\myparagraph{Annotator Disagreement}
Various works have faced the challenge of annotator disagreement for reasoning tasks. AmbiFC~\cite{glockner-etal-2021-ambifc}, concerned with whether a piece of evidence supports a given claim, explores the common factors causing disagreements. \citet{pavlick-kwiatkowski-2019-inherent} found persistent patterns of disagreement amongst RTE judgments not simply due to noise; efforts such as the UNLI protocol~\cite{chen-etal-2020-uncertain} aim to address this concern by modeling annotations on a logistic probability scale.

\myparagraph{Computational Argumentation}
Existing work has explored methods for computational argumentation; one subfield is argumentation mining~\cite{palau-etal-2009-argumentation}, which aims to detect and anaylze the arguments in a passage.
\citet{jin-etal-2022-logical} collect a dataset of items exhibiting 14 types of fallacies. %
\citet{stab-gurevych-2017-recognizing} hand-annotate the sufficiency of argumentative essays using RAS criteria, but are not geared towards entailment.

\myparagraph{Assessing Explanations}
Our work builds on literature evaluating the role and quality of model-generated explanations~\cite{wiegreffe-marasovic-2021-teach, tan-2022-diversity}. We add to this discourse by investigating how to improve the quality of decompositional entailment items that drive reasoning for complex tasks.
Similar to ours is the EEV methodology~\cite{valentino-etal-2021-natural}, under which an explanation is translated into logical form then assessed by a trained semanticist.
RDTE requires neither the translation nor the trained semanticist.

\section{Conclusion}

The trustworthy application of LLMs to complex reasoning tasks critically depends not only on accurate responses, but also accurate justifications. To date, research in entailment tree reasoning has largely focused on measuring the former.

This work develops a protocol based on works in informal logic for the assessment of compositional entailment, the core reasoning task undergirding entailment trees.  
We employed this rubric-based protocol to annotate RDTE, a high quality dataset of judged compositional entailments, along with tens of thousands of automatically scored items by GPT-4 under the same instructions, in multiple domains. 
We demonstrated that RDTE supports state-of-the-art results by a novel system for evidence-grounded entailment tree generation. 

The combination of our rubric, manual annotations, GPT-4 derived data, and this new system that we call \sysname{} represents a significant advance in building trustworthy AI systems capable of not only solving complex reasoning tasks, but providing correct justifications along with their answers.
\section{Limitations}
The \dataset{} dataset is a high-quality set of 1000 decompositions across two specific QA domains. As argument sufficiency is a domain-dependent notion, we had extensive discussion about what constituted validity in the two different tasks. Applying the RDTE protocol to new domains will likely also merit careful consideration of how the various facets of the task manifest differently for different types of questions.

A system such as \sysname{} does not carry direct risks towards others; however, since most automated reasoning systems can exacerbate biases already existing within language and culture, we recognize that our reasoning algorithm has inherent potential to cause damage to certain groups and identities.

\bibliography{anthology,custom}
\bibliographystyle{acl_natbib}

\appendix

\begin{table*}
\small
\setlength{\tabcolsep}{5pt}
\begin{tabular}{ll|rrr|rrr|rrr}
\toprule
 & & \multicolumn{3}{c}{\textbf{RDTE–ARC}} & \multicolumn{3}{c}{\textbf{RDTE–Hotpot}} & \multicolumn{3}{c}{\textbf{BaRDa}} \\ \cmidrule(lr){3-5} \cmidrule(lr){6-8}  \cmidrule(lr){9-11}  
 & \textbf{Training Data} & \textbf{Pr} & \textbf{Re} & \textbf{F$_{0.5}$} & \textbf{Pr} & \textbf{Re} & \textbf{F$_{0.5}$} & \textbf{Pr} & \textbf{Re} & \textbf{F$_{0.5}$} \\
\midrule
\multicolumn{4}{l}{\textbf{Itemwise GPT Methods}}  \\
GPT-4 (ICL) & RDTE Directions + 4 Exemplars & 55 & 90 & \textbf{59} & \textbf{49} & 74 & \textbf{53} & \textbf{92} & 46 & 76 \\
\quad (w/ Threshold 4) & & 49 & \textbf{100} & 55 & 45 & 86 & 50 & 90 & 58 & 81 \\
GPT-4 (Zero-Shot) & RDTE Directions Only  & 59 & 57 & 58 & 47 & 60 & 49 & 91 & 31 & 66\\
\quad (w/ Threshold 4) & & 39 & \textbf{100} & 45 & 35 & 91 & 40 & 83 & 70 & 80 \\
GPT-4 (BaRDa) & BaRDa Directions + 10 Exemplars & 40 & 72 & 44 & 43 & 95 & 48 & 82 & 85 & \textbf{83} \\
ChatGPT (ICL) & RDTE Directions + 4 Exemplars & 31 & 97 &  36 & 30 & 95 & 35 & 69 & 91 & 72 \\
ChatGPT (Zero-Shot) & RDTE Directions Only & 36 & 64 & 40 & 39 & 33 & \dn 38 & 72 & 15 & 41 \\
ChatGPT (BaRDa) & BaRDa Directions + 10 Exemplars  & 38 & 94 & \dn 43 & 30 & 79 & 34 & 73 & 84 & 75 \\  \midrule
\textbf{Batched GPT Methods} \\
GPT-4 (ICL) & & 52 & 79 & 56 & 52 & 62 & 54 & \multicolumn{3}{c}{(N/A)} \\
GPT-4 (Zero-Shot) & & 52  & 64 & 54 & 52 & 51 & 52 & \multicolumn{3}{c}{(N/A)}\\
\multicolumn{4}{l}{\textbf{T5 and Cross Encoders}}  \\
Entailer-11B & EntailmentBank + Entailer & 45 & 68 & 48 & 33 & 90 & 38 & 76 & 83 & 77 \\
NELLIE-3B & EntailmentBank + Entailer + QASC & 39 & 74 & 43 & 31 & 95 & 36 & 72 & 94 & 75 \\
NELLIE RoBERTa Filter & EntailmentBank + Entailer + QASC & 33 & 76 & 37 & 32 & 95 & 36 & 68 & \textbf{95} & 72 \\
OTS NLI RoBERTa & Various NLI incl QASC & 42 & 68 & \dn 45 & 39 & 85 & \dn 44 & 76 & 90 & 79  \\ \midrule 
\multicolumn{4}{l}{\textbf{Knowledge Distillation Students}} \\  
ChatGPT & GPT-4 (Zero-Shot) Silver Data & 46 & 58 & \up 48 & 52 & 49 & \up 51  & 82 & 55 & 75 \\
RoBERTa & GPT-4 (Zero-Shot) Silver Data & \textbf{68} & 57 & \up \textbf{66} & \textbf{56} & 56 & \up \textbf{56} & 83 & 47 & 72 \\
\bottomrule
\end{tabular}
\caption{Entailment results by various approaches on the \dataset{} and BaRDa datasets. Batching decompositions by their shared hypothesis does not drastically impact performance by GPT-4. Batching was not possible for BaRDa, which has one item per hypothesis. }
\label{tab:rdte_full_appendix}
\end{table*}

\section{Contrasting \nellie{} with \sysname{}}
\label{app:nellie_vs_treewise}

\nellie{}~\cite{weir-etal-2023-nellie} is a T5-based compositional entailment engine that shows high performance on Science QA by checking whether answer hypotheses are compositionally entailed by the WorldTree knowledge base~\cite{xie-etal-2020-worldtree}. Its search algorithm suffers from the primary drawbacks that (1) it requires a clean corpus of knowledge sentences, which is not always available for a particular problem domain, and (2) its various modules all rely on different in-domain training datasets not typically available for new tasks. 

In this section, we introduce \sysname{}, an evolution of \nellie{} based on instruction-tuned LLMs like ChatGPT and GPT-4. 
\sysname{} introduces a series of improvements to the engine's search algorithm that allow it to handle novel domains (1) with noisier knowledge sources like an index over Wikipedia passages common to state-of-the-art question-answering systems and (2) without module-specific training data. In this way, \sysname{} can answer whether a hypothesis from a novel QA dataset is compositionally entailed by Wikipedia.

\subsection{Search Logic}
We refer readers to \citet{weir-etal-2023-nellie} for an overview of the original \nellie{} search algorithm for compositionally grounding a hypothesis in a corpus. Their search generally follows a breadth-first search across candidate decompositions by following 3 Prolog rules:\footnote{We postulate that Prolog terms are evaluated in the sequence they are read, as is typical in most executors.}

\begin{enumerate}[leftmargin=.6cm]
\item[1.] \textit{Fact Unification} \\ $\scalemath{.85}{\textsc{Prove}(h) \Leftarrow \textsc{Retrieve}(h^{+}, f^{-}) \land \textsc{Entails}(f, h)}$   
    
\item[2.] \textit{Two Premise Rule Generation} \\ $\scalemath{.85}{\textsc{Prove}(h)\Leftarrow  \textsc{RuleGen}(h^+, f_1^-, f_2^-) \  \land} \\ 
\scalemath{.85}{\quad \textsc{Entails}( [f_1, f_2], h)}$ %
$\scalemath{.85}{\land \ \textsc{Prove}(f_1) \land \textsc{Prove}(f_2)}$ 

\item[3.] \textit{Retrieved First Premise Rule Generation} \\ $\scalemath{.85}{\textsc{Prove}(h)\Leftarrow 
\textsc{Retrieve}(h^{+}, f_1^{-}) \ \land} \\ \scalemath{.85}{\quad \textsc{RuleGen}(h^+, f_1^+, f_2^-)} 
\scalemath{.85}{\ \land \ \textsc{Entails}( [f_1, f_2], h)  \ \land \ }  \\ 
\scalemath{.85}{\quad \textsc{Prove}(f_2)}$
\end{enumerate}

We make the following observations:
\begin{enumerate}[label=(\alph*)]
    \item Rules 2/3 constrain the search to only binary conjunctions; allowing 3 or 4 might allow for added flexibility.
    \item The LM-calling \textsc{RuleGen} predicate only ever accepts 0 or 1 support facts; conditioning on multiple retrieved candidate facts might produce higher quality and/or more groundable decompositions.
    \item Rule 3 assumes that items returned by \textsc{Retrieve} ($f_1^{-}$) are of the same type as the premises the constitute an entailment: in \nellie{}'s case, simple evidential sentences like ``birds can fly.'' If \sysname{}'s corpus contains noisier knowledge, e.g. Wikipedia paragraphs, then this assumption becomes problematic.
    \item Rules 2/3 assume that any unique $f_1$, $f_2$, or $h$ is semantically distinct from other statements considered during the search. This implies that recursively calling $\textsc{Prove}$ on a new statement is never a waste of time. In practice, however, \nellie{} frequently considers hypotheses that are paraphrases of each other. This risks wasting substantial search time, especially if neither the hypothesis nor its paraphrase will ever be grounded.
\end{enumerate}

To address these issues, we make the following modifications, replacing rules 2 and 3 with rules 4, 2$^*$, 3$^*$. The latter two rules are only executed if 4 does not succeed. Bolded symbols denote string lists.
\begin{itemize}
    \item[4.] \textit{Paraphrase Unification} \\ $\scalemath{.85}{\textsc{Prove}(h) \Leftarrow \textsc{Expanded}(g^{-}) \ \land} $ \\
    $\scalemath{.85}{\quad  \textsc{Paraphrase}(h^{+}, g^{+}) \land \textsc{Prove}(g)}$
    \item[2$^\ast$.] \textit{Non-Conditioned Rule Generation} \\
    $\scalemath{.85}{\textsc{Prove}(h)\Leftarrow  \textsc{RuleGen}(h^+, [], \mathbf{f}) \  \land} \\ 
\scalemath{.85}{\quad \textsc{Entails}( \mathbf{f}, h)}$ 
$\scalemath{.85}{\land \ \textsc{MapList}(\mathbf{f}^+, \textsc{Prove})}$\footnote{\url{https://www.swi-prolog.org/pldoc/man?predicate=maplist/2}} 
    \item[3$^\ast$.] \textit{Retrieval-Conditioned Rule Generation} \\
    $\scalemath{.85}{\textsc{Prove}(h)\Leftarrow 
\textsc{Retrieve}(h^{+}, \mathbf{s}^{-}) \ \land} \\ 
\scalemath{.85}{
\quad \textsc{InferenceGen}(h^{+}, \mathbf{s}^{+}, \mathbf{i}^{-}) \ \land} \\ 
\scalemath{.85}{\quad \textsc{RuleGen}(h^+, \mathbf{i}^+, \mathbf{f}^-) \ \land \ \textsc{Entails}( \mathbf{f}, h) \ \land}$ \\
$\scalemath{.85}{\quad  \textsc{MapList}(\mathbf{f}^+, \textsc{Prove})}$
\end{itemize}

The novel predicate \textsc{Expanded} is a nondeterministic function that evaluates to true iff $g^-$ was a previous input to \textsc{RuleGen} during the search. The predicate \textsc{Paraphrase} uses SBERT~\cite{reimers-gurevych-2019-sentence} cosine similarity to identify
paraphrases. 
The predicate \textsc{InferenceGen} is a \textbf{forward chaining} inference generator that receives a list of support items (facts, passages, or otherwise) and returns a list of sentential premises likely entailed by the items that might be helpful to prove $h$. The new \textsc{RuleGen} now accepts an arbitrary list of candidate premises ($\mathbf{i}$) and returns an arbitrary-length decomposition ($\mathbf{f}$). As a result, the premises in the decomposition might not have appeared in $\mathbf{i}$ or $\mathbf{s}$; this adds flexibility to the generator but also means that the algorithm has to call \textsc{Prove} on all generated premises. If $f \in \mathbf{f}$ does appear in $\mathbf{i}$ or $\mathbf{s}$, it is likely immediately grounded via fact unification (rule 1).

\subsection{Prompt-based Modules}
\label{sec:prompts}
With the introduction of instruction-tuned LLMs like ChatGPT and GPT-4, we find it no longer necessary to train certain \nellie{} models via supervised learning on in-domain datasets. We replace the modules for query declarativization~\cite{demszky-etal-2018-transforming}, decomposition generation, and 1- and multi-premise entailment filtering with a mixture of in-context learning and zero-shot instruction prompts to a GPT model. This includes the predicates \textsc{RuleGen} and \textsc{Entails} above.
All prompts can be found in the appendix and will be released with the rest of the codebase.

\subsection{Improved Reasoning Generators}
A key improvement of \sysname{} over its predecessor is the redesigned approach to generating and filtering candidate decompositions for decompositional entailment.
\nellie{}'s original decomposition generator is a T5-based model trained via supervised learning on data from a specific domain. It is difficult to (A) adapt to new domains without retraining and (B) convey to the model that decompositions serve a reasoning-related purpose. With the introduction of instruction-tuned LLMs, this becomes more straightforward. For \sysname{}, we replace the T5-based generator with a diverse series of prompts for generating ad-hoc decompositions of a hypothesis in any domain:

\begin{itemize}
	\item A \textbf{fact-conditioned} prompt (\autoref{fig:fact_generator_prompt}) that receives a list of forward-chaining inferences derived from support documents using a separate prompt (\autoref{fig:forward_chain_prompt}) and returns a list of candidate decompositions.
	\item A \textbf{follow-up generation} prompt that receives the output of the previous prompt and an instruction to revise them to better fit the given hypothesis and question
	\item An \textbf{in-context learning} prompt (\autoref{fig:icl_generator_prompt}) dynamically constructed by retrieving exemplars from a fixed set of training items using BM25. We use as our training set the gold-annotated inferences from EntailmentBank~\cite{dalvi-etal-2021-explaining}.
\end{itemize}

Together these prompts populate an initial horizon of candidate decompositions to be condensed, checked for semantic equivalence, and subsequently filtered for argument validity.

\subsection{Reasoning Filters}
Our RDTE-oriented prompting strategy discussed in the main body of the paper is shown in \autoref{fig:icl_filter_prompt_1}, \ref{fig:icl_filter_prompt_2}, and \ref{fig:icl_filter_prompt_3}. The zero-shot variant contains the same directions but no exemplars. We use a separate set of exemplars for Hotpot vs ARC.

In addition to this improvement at multi-premise compositional NLI, we also implement a single-premise/passage entailment rubric for use by \sysname{} and when computing our tree integrity score. This prompt rates entailment on an ordinal 1-5 scale analogous to the RDTE protocol for compositional entailment; the prompt is shown in \autoref{fig:passage_entailment_prompt}.

\section{RDTE Annotation Instructions}
\label{app:ras_directions}
\autoref{fig:facet_instructions} shows the rubric and condition lists for evaluating premise-specific facts (relevance, factuality, redundancy). \autoref{fig:sufficiency_instructions} shows the rubric and condition list for annotating sufficency. These are ARC-specific lists; we constructed a similar but slightly different version for HotpotQA. We will release both rubrics with the dataset.

\section{Example \sysname{} Outputs}
\begin{figure*}[t!]
\centering
\footnotesize
\begin{tabular}{m{6.5cm} m{8.5cm}}
\toprule 
A balloon filled with water is placed in a freezer. Which property of the water will change as the water reaches its freezing point? (A) color, (B) mass, \textbf{(C) state}, (D) weight &
\multirow{4}{*}{\parbox{8.5cm}{
Which of these is not an instinctive behavior? (A) a bird building a nest, (B) a turtle burying its eggs, (C) a bear hibernating in winter, (D) a horse pulling a plow \\
\includegraphics[width=8.5cm]{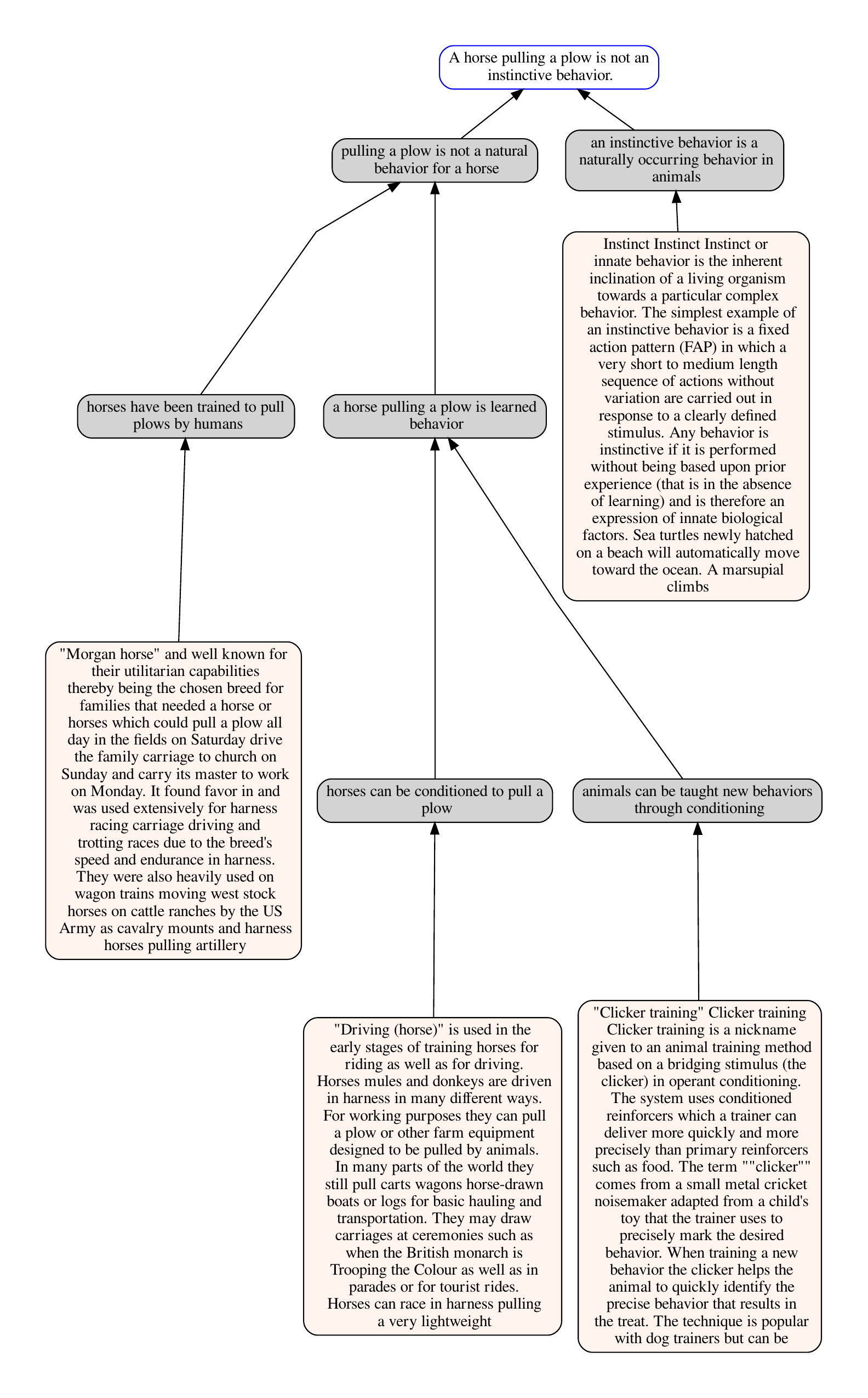}}} \\
\includegraphics[width=6.5cm]{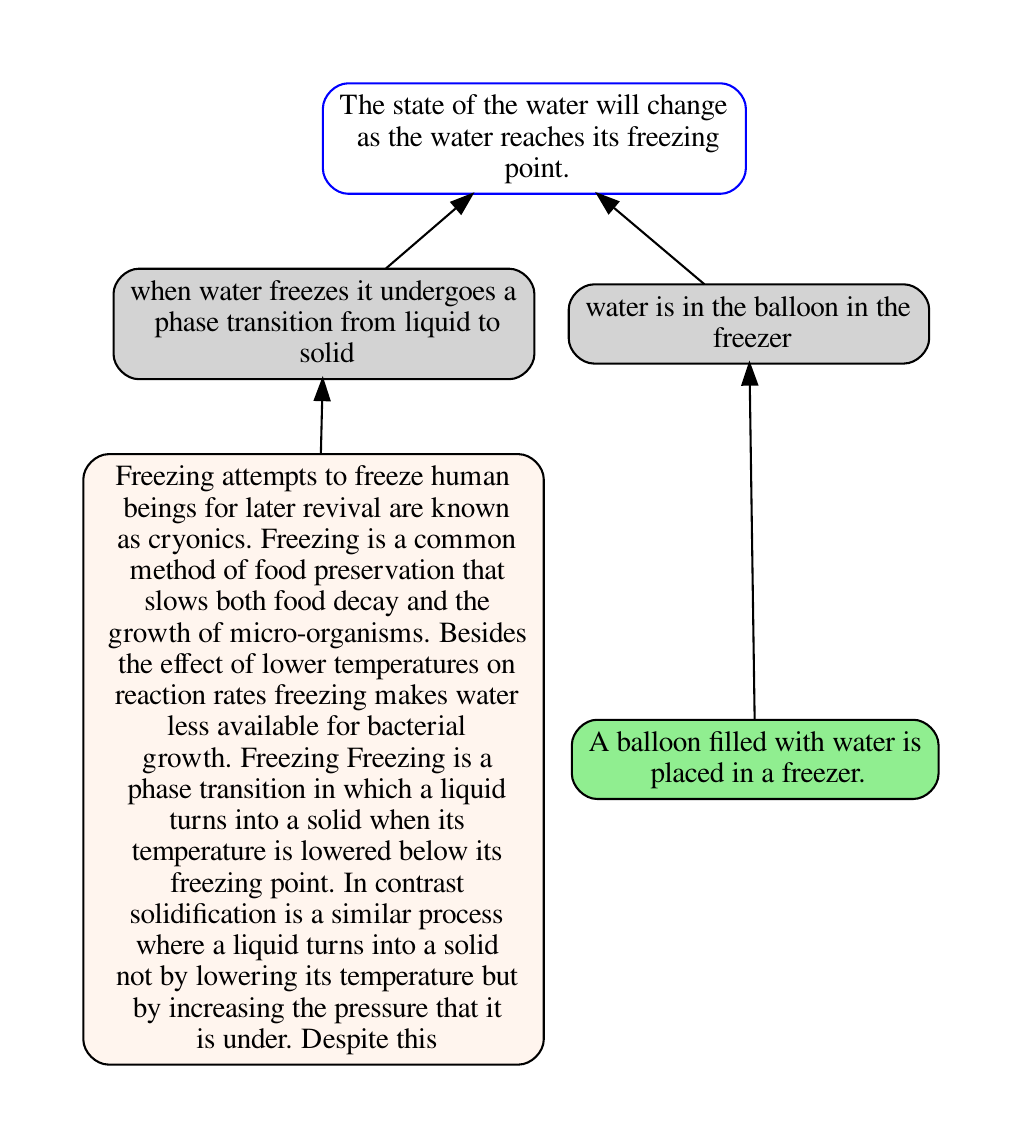} & \\

A laser beam is aimed at four different objects. Through which of these objects will the laser beam pass and be refracted? (A) a black cloth, (B) a piece of aluminum, (C) a sheet of paper, \textbf{(D) a glass prism} & \\
\includegraphics[width=7cm]{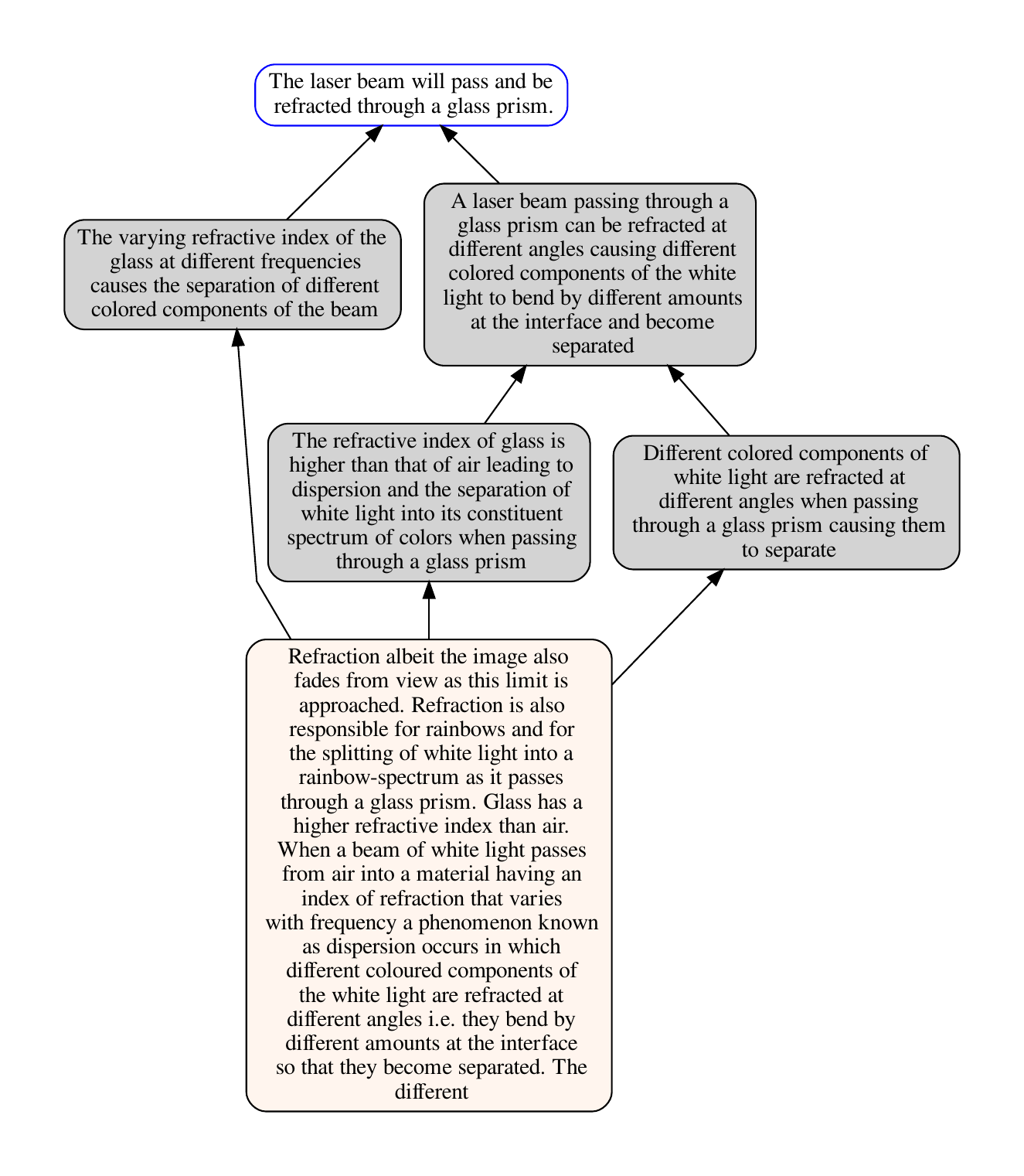} & \\
\bottomrule
\end{tabular}
\caption{Example multiple-choice questions from ARC with \sysname{}'s answer and corresponding proof grounded in Wikipedia.}
\label{fig:more-trees-arc}
\end{figure*}

\begin{figure*}
\centering
\footnotesize
\begin{tabular}{m{15cm}}
\toprule 
    Who was the New York City Fire Commissioner at the time of Providenza Panno's death? (A) Nicholas Scoppetta, (B) John J. Scannell, (C) William K. King, (D) Albert M. Arroyo, (E) Charles A. La Guardia, (F) Rhinelander Waldo, (G) James E. Langdon \\
    \includegraphics[width=11cm]{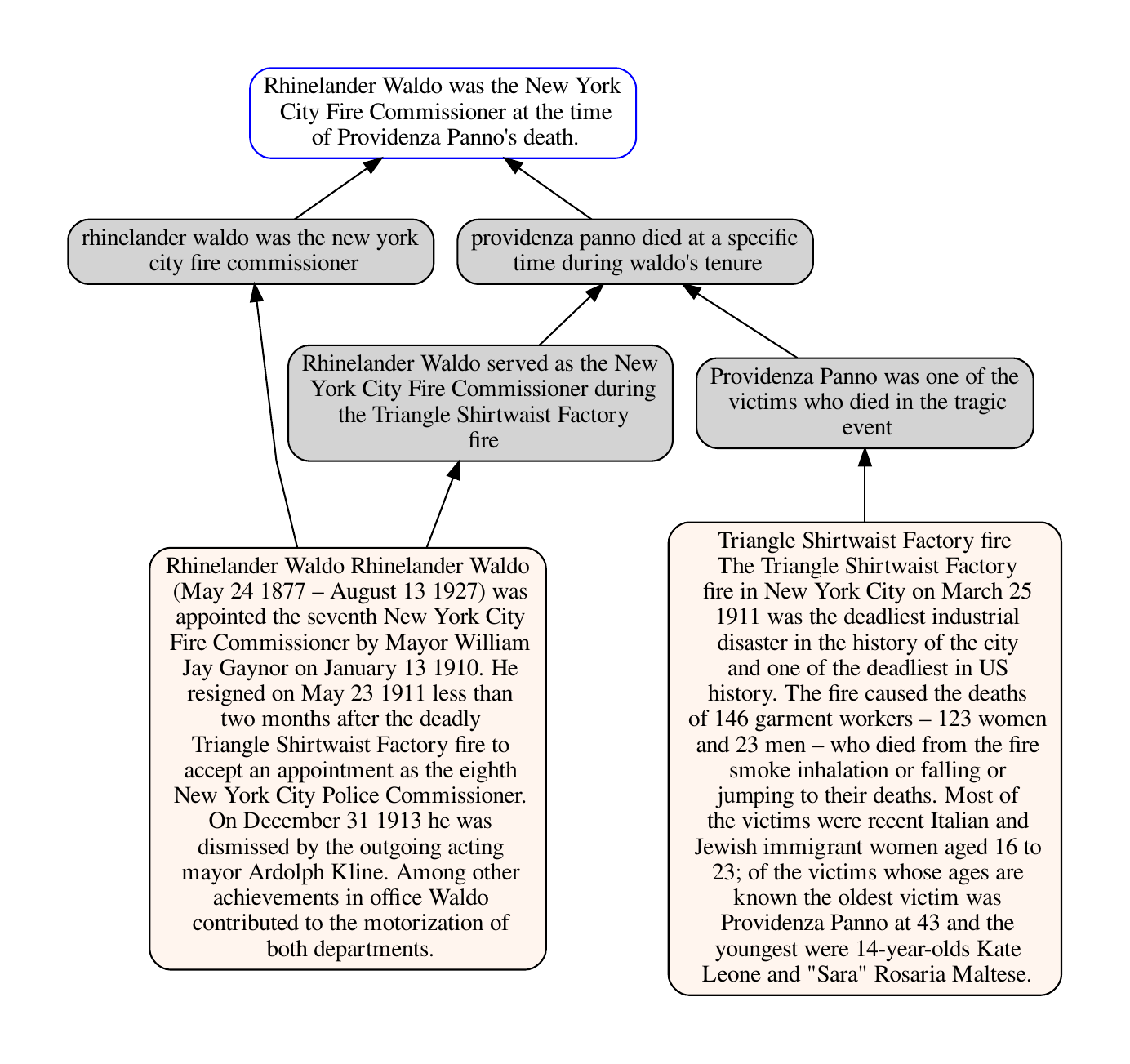}\\
    \midrule 
    Which Air Force member was behind enemy lines for 11 1/2 days and had the largest,longest and most complex rescue mission? \\
    \includegraphics[width=13cm]{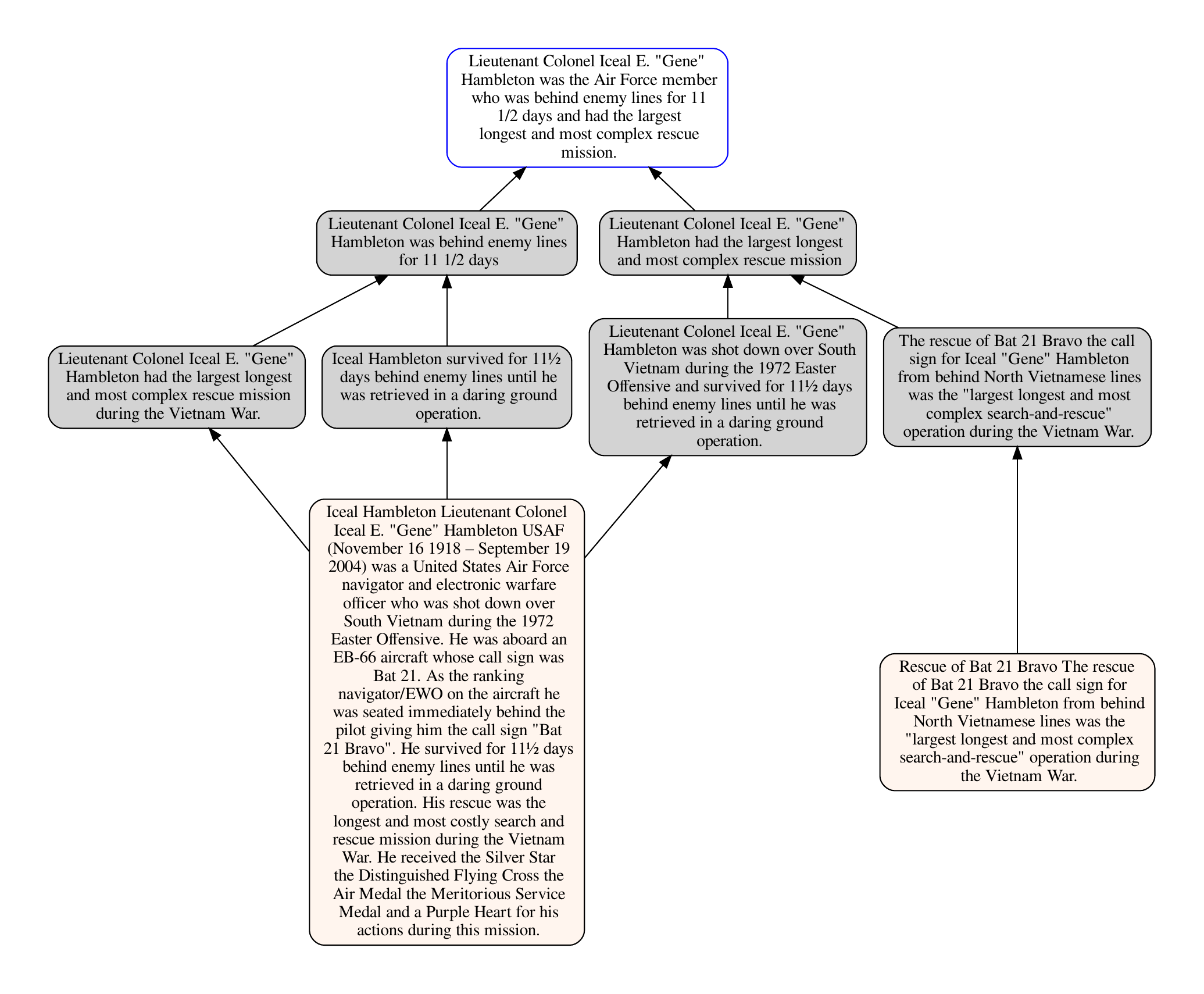} \\
    \bottomrule
\end{tabular}
\caption{Example multiple-choice questions from HotpotQA with \sysname{}'s answer and corresponding proof grounded in Wikipedia.}
\label{fig:more-trees-hotpot}
\end{figure*}

\autoref{fig:more-trees-arc} and \autoref{fig:more-trees-hotpot} depict example outputs by the \sysname{} tree search algorithm when hooked up to Wikipedia as the knowledge source for answering ARC and HotpotQA questions.

\begin{figure*}
\centering 
\footnotesize
\begin{tabular}{>{\bfseries}l p{12cm}}
\toprule
Factuality & How factual is the fact? 1 is completely false, 5 is completely true. \\ \midrule
Relevancy & How relevant is the fact to helping decompose the conclusion? An irrelevant fact is one that either does not address a key aspect of the conclusion, introduces irrelevant information, or is otherwise unnecessary should be scored lower in relevance. 1 is completely irrelevant, 5 is completely relevant. If the fact is situationally relevant to the conclusion, but contradicts it, you should still score it 5. \\ \midrule
Redundancy & Does the fact introduce new information that is not already contained in other facts in the decomposition? 1 is completely redundant, e.g., the third fact completely restates the first one, and 5 is completely new information. Sometimes, one of the facts directly restates the entire conclusion by itself, which \textbf{should be marked with the checkmark only and should not affect the numerical score}. Otherwise, facts including information included in the conclusion are acceptable and strictly necessary. \\
\bottomrule
\end{tabular}
\vspace{1cm}

\begin{tabular}{p{9.5cm}|p{4.8cm}} \toprule
\multicolumn{2}{l}{\textbf{{Factuality Questions to Ask Yourself}}} \\ \midrule
Is a fact ambiguously grounded in the question context in a way that does \textit{not} affect the reasoning? (e.g. the fact ``two sticks are rubbed together'' in the question ``what is an example of a force producing heat? (A) two sticks rubbed together, \dots '') & This is acceptable and can be 5/5 factuality \\ \midrule
Is a fact true in nearly all cases except extreme ones that don't pertain to the question? & \rfive (Yes = 5 factuality)\\
\midrule \midrule
\multicolumn{2}{l}{\textbf{Relevancy Questions to Ask Yourself}} \\ \midrule 
Is a fact not on topic? (``on topic'' is defined as containing nouns or entities that appear in the hypothesis) & \rone (Yes = 1 relevancy)  \\ \midrule
Does there not exist some (potentially over-pedantic) decomposition in which the given fact would be necessary to complete the entailment? & \rtwo (Yes = max 2 relevancy) \\ \midrule
Would removing an on-topic fact \textit{\textbf{in isolation}} \textbf{not} change the extent to which the conclusion is supported?   & \rtwo (Yes = 2 relevancy) \\ \midrule
Would removing an on-topic fact in isolation \textbf{minimally} change the extent to which the conclusion is supported?   & \rthree (Yes = 3 relevancy) \\ \midrule \midrule
\multicolumn{2}{l}{\textbf{Redundancy Questions to Ask Yourself}} \\ \midrule
Is a fact a paraphrase of another fact? & \rone (Yes = 1 redundancy\newline for second fact) \\ \midrule
Does a given fact add entailment in isolation, but if you removed the fact \textbf{\textit{conditioned on the rest of the facts}}, it \textbf{would not} change the extent to which the conclusion is supported?     & \rtwo (Yes = max 2 redundancy) \\ \midrule
Does a fact restate information in the \textbf{question text} (\textit{not} the conclusion)? & This is acceptable. Only check for restatements of other facts and/or the conclusion. Restatement of the question text, especially to cite evidence, is fine. \\ 
\bottomrule
\end{tabular}
\caption{\dataset{} annotation guidelines for premise-specific qualia in ARC.}
\label{fig:facet_instructions}
\end{figure*}
\begin{figure*}
\centering 
\footnotesize
\scalebox{.95}{
\begin{tabular}{>{\bfseries}p{3cm} p{11cm}}
\toprule
1 (Malformed or \newline Nonsensical) & Completely incorrect logic, or contains a fact that contradicts the conclusion, or malformed facts (not complete sentences), or inter-fact pronoun references (e.g. ``this'' or ``that'' or ``such''). \\ \midrule
2 (Poor) & Any two of the following: (1) some nontrivial amount of redundancy, (2) one irrelevant fact (2/5 or lower), (3) missing/implicit information that makes deducing the conclusion impossible without a substantial leap in logic. Would not convince a human of the conclusion. \\ \midrule
3 (Moderately \newline Correct) & Generally coherent and correct, but there is some significant flaw. E.g., \textbf{one of the facts is untrue} but if it was true the proof would be correct. \\ \midrule
4 (Mostly \newline Correct) & Slight redundancy or missing/implicit information, but \textbf{not to the point that it should substantially impact a human} performing the reasoning. \\ \midrule
5 (Perfect) & Completely correct and sound decomposition. No redundancy and no missing/implicit information. No ambiguous language. Addresses all conditions required to infer the conclusion. Does not leave anything implicit. \\
\bottomrule
\end{tabular}
}

\vspace{3mm}

\footnotesize 

\scalebox{.95}{
\begin{tabular}{p{12cm}|p{2.5cm}} \toprule
\multicolumn{2}{l}{\textbf{Questions to Ask Yourself}} \\ \midrule
Are any of the premises not well-formed? (no fragments, no 1 sentence that was split into two non-sentence parts) & \rone (Yes = 1) \\ \midrule
Do the premises together or individually \textit{contradict} the conclusion instead of supporting it? & \rone (Yes = 1)\\ \midrule
Are there between-premise pronoun references (`this', `that', `such') whose antecedent would be ambiguous without the other premises?  & \rone (Yes = 1) \\  \midrule 
Are there any conjunctive adverbs like ``therefore" or ``thus"? & \rone (Yes = 1) \\  \midrule 
Are all premises irrelevant, off-topic, or not contributing any correct logic? (e.g. all 1's for relevance, or removing all of them would not change the extent of the entailment) & \rone (Yes = 1) \\ \midrule 
Does any premise essentially restate the conclusion without adding/removing any information? & \rtwo (Yes = max 2) \\ \midrule
Are there at least two of the following? (1) redundant fact, (2) untrue fact, (3) irrelevant fact, (4) missing information & \rtwo (Yes = max 2) \\ \midrule 
Is the conclusion assuming an effect that isn't directly linked to the cause, or overlooking more immediate effects? & \rtwo (Yes = max 2) \\ \midrule
If you removed all redundant or irrelevant (3/5 or lower) facts, would there be only one fact remaining \textbf{and not full entailment}? & \rtwo (Yes = max 2) \\ \midrule
Is there any amount of logical error present? & \rtwo (Yes = max 2) \\ \midrule 
Did you give any fact a 2/5 or lower for factuality or relevance? & \rthree (Yes = max 3) \\ \midrule
Does proving one premise amount to proving all of the others? & \rthree (Yes = max 3) \\ \midrule
Are the premises all factual and relevant, but there is a part of the conclusion (e.g. something non-common-sense or a thing that a 10-year-old would \textbf{not} intuit in the context of the question) that is not stated or explained? & \rthree (Yes = max 3) \\ \midrule 
If you removed all redundant or irrelevant (3/5 or lower) facts, would there be only one fact remaining? & \rthree (Yes = max 3) \\ \midrule
Are the premises all evidence statements entailed by the question context and nothing else? & \rthree (Yes = \textbf{likely} max 3) \\ \midrule
Is one of the facts not true, but if it were then it'd be a perfect entailment? & \rthree (Yes = 3) \\ \midrule
Did you give any fact a 3/5 or lower for redundancy, factuality, or relevance? & \rfour (Yes = max 4) \\ \midrule
Are there two separate arguments being partially/mostly made to support the hypothesis, but one/both is missing some implicit premises? & \rfour (Yes = max 4) \\ \midrule
Are the premises all factual and relevant, but the conclusion does not follow from the premises for a minor reason (e.g. a common sense-y fact that would have been inferred by a 10-year-old in the context of the question) & \rfour (Yes = max 4) \\ \midrule
Do two of the facts perfectly entail the conclusion, but the third is essentially redundant? & \rfour (Yes = 4)\\ \midrule
Is the conclusion irrelevant to the question, but the entailment supports the conclusion perfectly? & \rfive (Yes = 5)\\ \midrule
Does the question ask something along the lines of ``which is the best\dots?'' and the entailment doesn't mention the other answer options? & Treat it like the ``best'' is not there \\  \midrule 
Do the premises not properly entail the conclusion for some other reason? & Reach out to us for clarification  \\ \midrule 
Is one premise P1 an effective paraphrase of the hypothesis H, but another premise P2 serves as lexical grounding between P1 and H? & Ignore the paraphrase = max 2 rule \\
\bottomrule
\end{tabular}
}
\caption{\dataset{} annotation guidelines for annotating sufficiency in ARC.}
\label{fig:sufficiency_instructions}
\end{figure*}

\begin{figure*}

\paragraph{RDTE judgment prompt:} 
\quad \\
\vspace{-5mm}
\begin{lstlisting}[style=base]
You are a reasoning system that searches for proofs of a hypothesis by recursively decomposing it into simpler premises.  
  
Given a question and a hypothesis, you give a list of possible decompositions of the hypothesis into premises such that proving the list of premises would amount to proving the hypothesis through compositional entailment. The hypothesis might represent an answer to the question ,or it might represent a recursive query. However, many of the decompositions are incorrect, and you must identify which ones are correct and which ones are incorrect. For the following question, hypothesis and list of premise decompositions, score each decomposition according to the following rubrics:  
  
You will first score each premise on a scale of 1 to 5 for each of the following qualia:  
Factuality: How factual is the premise? 1 is completely false, 5 is completely true.  
Relevance: How relevant is the premise to helping explain the hypothesis? 1 is completely irrelevant, 5 is completely relevant.  
Redundancy: Does the premise introduce new information that is not already contained in other premises? 1 is completely redundant, i.e. completely restating another premise or the hypothesis, 5 is completely new information.  
  
You will then judge whether the decomposition as a whole constitutes a complete inference, on a scale of 1 to 5 using the following rubric:  
1 (malformed or nonsensical): Completely incorrect logic, or contains a premise that contradicts the hypothesis, or malformed instances, or inter-premise pronoun references (E.g. a "this" in premise 2 that refers to premise 1).)  
2 (poor): Some nontrivial amount of redundancy, one irrelevant fact, and/or missing information. Would impact a human performing reasoning.  
3 (moderately correct): Generally coherent and correct, but there is some significant flaw. (E.g., one of the facts is untrue but if it was true the proof would be correct.)  
4 (mostly correct): Slight redundancy or missing information, but not to the point that it should substantially impact a human performing the reasoning.  
5 (perfect): Completely correct and sound decomposition. No redundancy and no missing information. No ambiguous language.  
  
You are renowned for your stringent eye. There should be minimal "information loss" between the hypothesis and the premises; you are looking for strict entailment.  YOU RARELY GIVE A 5  
  
Finally, you will provide an explanation for your judgment. Your explanation should justify any non-perfect scores you have given for factuality, relevance, and redundancy. In other words, explain why you gave a premise a certain score based on the information in the premise and its relation to the hypothesis.   
For the complete inference score, explain why the conjunction of premises either does or does not amount to a complete and correct proof of the hypothesis. If there were issues with the complete inference, identify what information was missing or what logical errors were made. Your explanation should be clear and concise, providing valuable insight into your scoring process  
  
Your output should be serialized json items, one per line, and nothing else.   
  
QUESTION 1:   
Which of the following items conducts electricity? (A) a lego brick, (B) a suit of armor, (C) a wooden table, (D) a T-shirt  
  
HYPOTHESIS 1:  
A suit of armor conducts electricity  
  
DECOMPOSITIONS 1:  
(1) a suit of armor is made of iron AND iron is a metal  
(2) armor is made of metal AND metal conducts electricity  
(3) armor cannot be punctured AND iron conducts electricity  
(4) armor is made of iron AND iron is a metal AND metal conducts electricity  
(5) armor is an object AND objects conduct electricity AND armor is an object that is made of metal  
(6) a wooden table is made of wood AND wood conducts electricity  
(7) conductivity is the degree to which a material conducts electricity AND conductivity is measured as the ratio of current density to the electric field that causes the flow of current.   
(8) an item conducts electricity if the material that it is made of conducts electricity AND metal conducts electricity  
(9) a suit of armor is made of iron AND iron is a metal AND conductivity is measured in Siemens per meter  
(10) metal conducts electricity AND a suit of armor conducts electricity  
  
JUDGEMENTS 1 (10 items):  
{{"index": 1, "factuality": [4, 5], "relevance": [5, 5],  "redundancy": [5, 5], "complete_inference": 2, "explanation": "The fact that armor is made of iron and iron is a metal does not necessarily mean that armor conducts electricity."}}  
{{"index": 2, "factuality": [5, 5], "relevance": [5, 5],  "redundancy": [5, 5], "complete_inference": 5, "explanation": "Properly identifies that armor is made of a type of material (metal) that conducts electricity."}}  
{{"index": 3, "factuality": [5, 5], "relevance": [1, 5],  "redundancy": [5, 5], "complete_inference": 2, "explanation": "The fact that armor cannot be punctured is irrelevant to whether armor conducts electricity."}}  
{{"index": 4, "factuality": [3, 5, 5], "relevance": [5, 5, 5],  "redundancy": [5, 5, 5], "complete_inference": 4, "explanation": "identifies that armor is made of a material (iron) that is a type (metal) that conducts electricity, but armor is not always made of iron."}}  
{{"index": 5, "factuality": [5, 2, 5], "relevance": [3, 3, 5],  "redundancy": [1, 5, 5], "complete_inference": 3, "explanation": "Not all objects conduct electricity, so the premise 'objects conduct electricity' is an overgeneralization."}}  
{{"index": 6, "factuality": [5, 1], "relevance": [1, 2],  "redundancy": [5, 5], "complete_inference": 2, "explanation": "The premises are about a wooden table, not referencing a suit of armor, and wood does not conduct electricity."}}  
{{"index": 7, "factuality": [5, 5], "relevance": [1, 1],  "redundancy": [5, 5], "complete_inference": 2, "explanation": "The premises are about the general concept of conductivity, not specifically about a suit of armor or metal."}}  
{{"index": 8, "factuality": [5, 5], "relevance": [5, 5],  "redundancy": [5, 5], "complete_inference": 3, "explanation": "Does not include that a suit of armor is made of metal"}}  
{{"index": 9, "factuality": [5, 5, 5], "relevance": [5, 5, 1],  "redundancy": [5, 5, 5], "complete_inference": 2, "explanation": "the measurement of conductivity in Siemens per meter is unnecessary to prove that a suit of armor conducts electricity."}}  
{{"index": 10, "factuality": [5, 5], "relevance": [5, 5],  "redundancy": [5, 1], "complete_inference": 2, "explanation": "Second premise restates the hypothesis"}}  
\end{lstlisting}
\vspace{-3mm}
\caption{Directions and in-context learning exemplar prompt used to extract RDTE judgments from instruction-tuned models. The zero-shot variant has the same first half but no examples.}

\label{fig:icl_filter_prompt_1}
\end{figure*}

\begin{figure*}
\paragraph{RDTE judgment prompt (continued):} 
\quad \\
\vspace{-5mm}
\begin{lstlisting}[style=base]
QUESTION 2:  
Natural selection cannot occur without (A) competition for unlimited resources. (B) gradual warming of Earth. (C) genetic variation in species. (D) asexual reproduction in species.  
  
HYPOTHESIS 2:  
Natural selection cannot occur without competition for unlimited resources.  
  
DECOMPOSITIONS 2:  
(1) Competition for resources can lead to individuals with favorable traits reproducing more AND Natural selection is the process where organisms with favorable traits are more likely to reproduce  
(2) Competition is a natural occurrence when resources are limited AND Unlimited resources can lead to an absence of competition AND Without competition, there's no natural selection  
(3) The limited availability of a required resource can make survival and growth difficult for a species AND if an organism survives competition for resources then this organism will survive / reproduce for natural selection  
(4) competition for resources is a selection pressure AND in nature, resources are limited AND natural selection occurs when there is competition for resources  
(5) competition for unlimited resources is necessary for natural selection to occur AND limited availability of a required resource can make survival and growth difficult for a species  
(6) natural selection requires competition for unlimited resources AND the survival of species depends on available resources and lack of competition  
(7) competition for unlimited resources is a form of competition that drives natural selection AND natural selection is the process by which organisms with beneficial traits survive competition to reproduce and pass on those traits  
(8) The survival of species depends on available resources and lack of competition AND limited availability of water can be a selection pressure leading to evolutionary adaptations AND without photosynthesis, a plant cannot produce the food it needs to grow and survive  
(9) Competition for unlimited resources could be treated as a driving pressure for evolution AND Natural selection is one of the mechanisms of evolution  
(10) competition for resources is a form of competition that drives natural selection AND natural selection is the process by which organisms with beneficial traits survive competition to reproduce and pass on those traits  
(11) competition for unlimited resources provides a greater opportunity for organisms with beneficial traits to have an advantage in survival and reproduction AND natural selection requires competition for resources in order for organisms with beneficial traits to have an advantage in survival and reproduction  
(12) The survival of species depends on available resources and lack of competition AND in nature, resources are limited  
  
JUDGEMENTS 2 (12 items):  
{{"index": 1, "factuality": [5, 5], "relevance": [5, 5],  "redundancy": [5, 5], "complete_inference": 3, "explanation": "The decomposition does not specify the role of 'unlimited resources' in the competition."}}  
{{"index": 2, "factuality": [5, 5, 5], "relevance": [5, 1, 5],  "redundancy": [5, 5, 5], "complete_inference": 2, "explanation": "The second premise weakens the hypotheses, the first premise is irrelevant, and the last premise doesn't address unlimited resources"}}  
{{"index": 3, "factuality": [5, 5], "relevance": [5, 5],  "redundancy": [5, 5], "complete_inference": 2, "explanation": "The decomposition does not specify the role of 'unlimited resources' in the competition."}}  
{{"index": 4, "factuality": [5, 5, 5], "relevance": [5, 2, 5],  "redundancy": [5, 5, 5], "complete_inference": 2, "explanation": "The decomposition does not specify the role of 'unlimited resources' in the competition, and premise 2 is not very relevant"}}  
{{"index": 5, "factuality": [5, 5], "relevance": [5, 5],  "redundancy": [1, 5], "complete_inference": 2, "explanation": "the first premise restates the hypothesis"}}  
{{"index": 6, "factuality": [5, 1], "relevance": [5, 5],  "redundancy": [5, 5], "complete_inference": 2, "explanation": "The second premise is untrue, the survival of species does not depend on a lack of competition."}}  
{{"index": 7, "factuality": [1, 5], "relevance": [5, 5],  "redundancy": [5, 5], "complete_inference": 3, "explanation": "premise 1 is false."}}  
{{"index": 8, "factuality": [1, 5, 5], "relevance": [1, 1, 1],  "redundancy": [5, 5, 5], "complete_inference": 1, "explanation": "The premises are about water and photosynthesis, not competition for unlimited resources."}}  
{{"index": 9, "factuality": [2, 5], "relevance": [5, 5],  "redundancy": [5, 5], "complete_inference": 2, "explanation": "premise 1 is effectively untrue, and the decomposition does not explain how natural selection requires the competition for resources."}}  
{{"index": 10, "factuality": [5, 5], "relevance": [5, 5],  "redundancy": [3, 5], "complete_inference": 4, "explanation": "Premise 1 is somewhat redundant given premise 2, and the decomposition does not clearly explain why unlimited resources specifically are necessary for natural selection to occur."}}  
{{"index": 11, "factuality": [1, 5], "relevance": [5, 5],  "redundancy": [5, 5], "complete_inference": 3, "explanation": "premise 1 is untrue."}}  
{{"index": 12, "factuality": [1, 5], "relevance": [1, 1],  "redundancy": [5, 5], "complete_inference": 2, "explanation": "The first premise is untrue, the survival of species does not depend on a lack of competition, and the second is irrelevant to the question of unlimited resources"}}  
  
QUESTION 3:  
The gravitational force between the Moon and Earth depends on (A) their masses, only, (B) their diameters, only, (C) their masses and how far apart they are, (D) their diameters and how far apart they are  
  
HYPOTHESIS 3 (RECURSIVE):  
the Moon and Earth are two objects  
  
DECOMPOSITIONS 3:  
(1) the Moon and Earth are at a certain distance apart AND the Moon and Earth have masses  
(2) Earth is an object AND planets are objects AND the Moon is an object  
(3) Earth is an object AND natural satellites are objects AND the Moon is a natural satellite  
(4) the Moon and Earth have diameters AND the Moon and Earth have masses  
(5) the Moon and Earth are at a certain distance apart AND the Moon and Earth have diameters  
(6) all existing entities in space are regarded as objects AND the Earth exists AND the Moon exists  
(7) the Earth is an object AND the Moon is an object AND two objects can exert gravitational force on each other  
(8) Earth is an object AND the Moon is an object  
(9) objects with mass are considered as objects in physics AND the Moon and Earth both have mass AND the Earth has mass  
(10) the Earth is an object AND the Moon is an object  
\end{lstlisting}
\vspace{-3mm}
\caption{In-context learning prompt for using the RDTE protocol (2/3)}
\label{fig:icl_filter_prompt_2}
\end{figure*}

\begin{figure*}
\paragraph{RDTE judgment prompt (continued):}  \quad \\ \vspace{-5mm}
\begin{lstlisting}[style=base]
JUDGEMENTS 3 (10 items):  
{{"index": 1, "factuality": [5, 5], "relevance": [1, 5],  "redundancy": [5, 5], "complete_inference": 2, "explanation": "premise 1 is irrelevant to the hypothesis"}}  
{{"index": 2, "factuality": [5, 5, 5], "relevance": [5, 5, 5],  "redundancy": [5, 2, 5], "complete_inference": 4, "explanation": "that planets are objects is unnecessary given premise 1"}}  
{{"index": 3, "factuality": [5, 5, 5], "relevance": [5, 5, 5],  "redundancy": [5, 5, 5], "complete_inference": 5, "explanation": "The premises correctly entail that the Earth and the Moon (a natural satellite) are both objects."}}  
{{"index": 4, "factuality": [5, 5], "relevance": [1, 5],  "redundancy": [5, 5], "complete_inference": 2, "explanation": "premise 1 is irrelevant"}}  
{{"index": 5, "factuality": [5, 5], "relevance": [1, 1],  "redundancy": [5, 5], "complete_inference": 1, "explanation": "both premises are irrelevant to the hypothesis"}}  
{{"index": 6, "factuality": [5, 5, 5], "relevance": [5, 5, 5],  "redundancy": [5, 5, 5], "complete_inference": 5, "explanation": "correctly identifies that the Earth and the Moon, which both exist, are considered objects in space."}}  
{{"index": 7, "factuality": [5, 5, 5], "relevance": [5, 5, 1],  "redundancy": [5, 5, 5], "complete_inference": 3, "explanation": "the third premise about gravitational force is not necessary to prove the hypothesis."}}  
{{"index": 8, "factuality": [5, 5], "relevance": [5, 5],  "redundancy": [5, 5], "complete_inference": 5, "explanation": "The premises properly entail that both things are objects."}}  
{{"index": 9, "factuality": [5, 5, 5], "relevance": [5, 5, 5],  "redundancy": [5, 5, 1], "complete_inference": 3, "explanation": "Premise 3 is redundant given premise 2"}}  
{{"index": 10, "factuality": [5, 5], "relevance": [5, 5],  "redundancy": [5, 5], "complete_inference": 5, "explanation": "directly states that both the Earth and the Moon are objects."}}  
  
QUESTION 4:  
{question}  
  
HYPOTHESIS 4 {recursive_or_not}:  
{hypothesis}  
  
DECOMPOSITIONS 4:  
{decompositions}  
  
JUDGEMENTS 4 ({n_items} items):
\end{lstlisting}
\vspace{-3mm}
\caption{In-context learning prompt for using the RDTE protocol (3/3) }
\label{fig:icl_filter_prompt_3}
\end{figure*}

\begin{figure*}
    \begin{lstlisting}[style=base]
In the following exercise, I would like you to tell me if a line of reasoning is reasonable or not.  
 
I will give you some facts and a possible conclusion. Please tell me whether the conclusion reasonably follows from the facts I gave you. If the conclusion does reasonably follow from the facts, then please answer "yes".  If the conclusion does not reasonably follow from the facts, then please answer "no".  
 
Note that some of the facts may be false, but I am only interested whether the conclusion would reasonably follow IF those facts were true. In other words, imagine a world in which the given facts are true. Would it be reasonable to draw the conclusion from those facts, if they were true?  
 
Here are some examples: 
 
IF Vegetables are plants.   
AND Cabbages are plants.   
THEN Cabbages are vegetables. 
Q: Does the rule's conclusion reasonably follow from the facts in the condition, if they were true? A: no 
 
IF a nail is made of metal 
AND metals conduct electricity 
THEN a nail conducts electricity. 
Q: Does the rule's conclusion reasonably follow from the facts in the condition, if they were true? A: yes 
 
IF dogs are birds 
AND birds can fly 
THEN dogs can fly 
Q: Does the rule's conclusion reasonably follow from the facts in the condition, if they were true? A: yes 
 
IF sound requires matter to travel   
AND a vacuum has no matter in it   
THEN sound will not travel in a vacuum. 
Q: Does the rule's conclusion reasonably follow from the facts in the condition, if they were true? A: yes 
 
IF Erosion can cause a landslide.   
AND Mud is deposited by a landslide.   
THEN Erosion can cause mud to be deposited. 
Q: Does the rule's conclusion reasonably follow from the facts in the condition, if they were true? A: yes 
 
IF An animal needs to breathe in order to live.   
AND Living things need water to live.   
THEN Animals need water to live. 
Q: Does the rule's conclusion reasonably follow from the facts in the condition, if they were true? A: yes 
 
IF Frogs also have a larynx, or voice box, to make sounds.  
AND Animals that have vocal cords can make sounds.  
THEN Frogs are animals. 
Q: Does the rule's conclusion reasonably follow from the facts in the condition, if they were true? A: no 
 
IF All humans breathe.   
AND Stones breathe. 
THEN All humans and stones breathe. 
Q: Does the rule's conclusion reasonably follow from the facts in the condition, if they were true? A: yes 
 
IF If a planet is rocky, it can only have a thin atmosphere.   
AND Small planets and rocky planets have very thin atmospheres.  
THEN If a planet is small and rocky, it has a thin atmosphere. 
Q: Does the rule's conclusion reasonably follow from the facts in the condition, if they were true? A: yes 
 
IF Damming a river can cause a lake to form.   
AND Dams are made of concrete.  
THEN Dams are concrete lakes.  
Q: Does the rule's conclusion reasonably follow from the facts in the condition, if they were true? A: no 
 
Now your turn!
{Entailment}
\end{lstlisting}
------
\begin{lstlisting}[style=base]
Answer the following yes/no question with either "yes" or "no". Just give a single word answer. Do not give any explanation or justification. 
 
Here are some examples: 
Is it true that an ocean contains large bodies of water? yes 
Is it true that lightning is similar to a volcano erupting? no 
Is it true that a fox squirrel is a kind of animal? yes 
Is it true that a rainbow is a kind of electromagnetic discharge? no 
Is it true that the surface of the moon is made up of water? no 
Is it true that the surface of the moon is made up of gases? no 
Is it true that a bluebird is a kind of animal? yes 
Is it true that the moon 's surface is made up of oceans? no 
Is it true that the opposite of negative impact is positive impact? yes 
Is it true that building a new highway through the area has a negative impact on the ecosystem? yes 
 
Now let's do some more! Remember, answer with just a single word, yes or no. 
Is it true that} \normalsize {\it insert the statement to assess here}
\end{lstlisting}
    \caption{BaRDa prompts from \citet{clark-etal-2023-barda} for entailment (upper) and factuality (lower) judgments.}
    \label{fig:barda-prompt}
\end{figure*}

\begin{figure*}

\paragraph{Document-conditioned forward generation prompt:} \quad \\ \vspace{-5mm}

\footnotesize
\begin{lstlisting}[style=base]
You are a reasoning system that searches for proofs of a hypothesis by decomposing into simpler premises. 

For the following hypothesis and source documents, write a set of independent inferences entailed by one or multiple documents. The inferences should resemble world facts and should help to decompose the hypothesis into component reasoning steps. The inferences should NOT simply restate the hypothesis. 

Your output should be a serialized json item, one per line, with the format {{"inference": <inference text>, "source": [<indices of source documents>]}} and nothing else. 

QUESTION: 
{question} 

HYPOTHESIS:  
{hypothesis} 

SOURCE DOCUMENTS YOU MIGHT PULL FROM: 
{documents}

{n} INFERENCES THAT MIGHT SUPPORT HYPOTHESIS: 
\end{lstlisting}

\caption{Prompt used for creating forward-chaining inferences from retrieved source documents}
\label{fig:forward_chain_prompt}

\end{figure*}
\begin{figure*}

\paragraph{Exemplar-conditioned decomposition generation prompt:} \quad \\ \vspace{-5mm}

\begin{lstlisting}[style=base]
You are a reasoning system that searches for proofs of a hypothesis by decomposing into simpler premises.

Given a question and corresponding hypothesis, you give a list of 20 possible decompositions of the hypothesis into two or three facts, F1 and F2 (and possibly F3), such that proving the list of Fs would amount to proving the hypothesis through compositional entailment. There should be minimal "information loss" between the hypothesis and the Fs; you are looking for strict entailment.

Each decomposition should be some combination of core scientific principles as well as conclusions about the question at hand. They should not imply each other, i.e. none of them should start with "thus" or "therefore".

You also optionally take a list of facts that you might use in your decompositions.

Your 20 decompositions should follow different "reasoning patterns." Try to create decompositions that are semantically distinct and make use of different core facts or underlying principles.

Your output should be a serialized json item, one per line, with the format {"fact1": <fact1>, "fact2": <fact2>, "fact3" : <fact3, if necessary>} and nothing else.



QUESTION:
An ecosystem is a community of organisms interacting with their physical environment. Why are decomposers an important part of ecosystems? (A) They break down dead organisms to return nutrients to the soil. (B) They produce their own food for survival. (C) They play a role in preventing weathering and erosion. (D) They provide most of the energy to consumers.

HYPOTHESIS:
Decomposers are an important part of ecosystems because they break down dead organisms to return nutrients to the soil.

4 DIFFERENT POSSIBLE DECOMPOSITIONS, 2 OR 3 FACTS EACH, ONE JSON ITEM PER LINE:
{"fact1": "a decomposer breaks down dead organisms to return nutrients to soil", "fact2": "nutrients in soil are important for an ecosystem"}
{"fact1": "decomposition is when a decomposer breaks down dead organisms", "fact2": "decomposition is when a decomposer recycles / returns nutrients / nitrogen from dead organisms to the soil by eating those dead organisms", "fact3": "nutrients in soil are important for an ecosystem"}
{"fact1": "a decomposer breaks down dead organisms to return nutrients to soil", "fact2": "nutrients in soil are important to plants", "fact3": "plants are a part of an ecosystem"}



QUESTION:
What is the role of decomposers in a food chain? (A) They consume other organisms. (B) They break down dead organic matter. (C) They use the Sun's energy to make food. (D) They convert inorganic matter into organic matter.

HYPOTHESIS:
The role of decomposers in a food chain is they break down dead organic matter.

2 DIFFERENT POSSIBLE DECOMPOSITIONS, 2 OR 3 FACTS EACH, ONE JSON ITEM PER LINE:
{"fact1": "an organism is a source of organic matter", "fact2": "decomposer is a kind of role in the food chain process / in an ecosystem", "fact3": "decomposition is when a decomposer breaks down dead organisms"}
{"fact1": "an organism is a source of organic matter", "fact2": "the role of decomposers in the food chain process is to break down dead organisms"}



QUESTION:
{question}

HYPOTHESIS:
{hypothesis}

20 DIFFERENT POSSIBLE DECOMPOSITIONS, 2 OR 3 FACTS EACH, ONE JSON ITEM PER LINE::

\end{lstlisting}

\caption{Prompt used for creating exemplar-conditioned decompositions. Exemplars are retrieved from the EntailmentBank training set using BM25 with the question and hypothesis as query.}
\label{fig:icl_generator_prompt}

\end{figure*}
\begin{figure*}

\paragraph{Fact-conditioned decomposition generation prompt:} \quad \\ \vspace{-5mm}

\begin{lstlisting}[style=base]
You are a reasoning system that searches for proofs of a hypothesis by decomposing into simpler premises. 

Given a question and corresponding hypothesis, you give a list of {n_candidates} possible decompositions of the hypothesis into two or three facts, F1 and F2 (and possibly F3), such that proving the list of Fs would amount to proving the hypothesis through compositional entailment. There should be minimal "information loss" between the hypothesis and the Fs; you are looking for strict entailment. 

Each decomposition should be some combination of core scientific principles as well as conclusions about the question at hand. They should not imply each other, i.e. none of them should start with "thus" or "therefore". 

You also optionally take a list of facts that you might use in your decompositions. 

Your {n_candidates} decompositions should follow different "reasoning patterns." Try to create decompositions that are semantically distinct and make use of different core facts or underlying principles. 

Your output should be a serialized json item, one per line, with the format {{"fact1": <fact1>, "fact2": <fact2>, "fact3" : <fact3, if necessary>}} and nothing else. 

QUESTION: 
{question} 

HYPOTHESIS: 
{hypothesis} 

FACTS YOU MIGHT USE, IF THEY ARE RELEVANT: 
{facts}  

{n_candidates} DIFFERENT POSSIBLE DECOMPOSITIONS, ONE JSON ITEM PER LINE:  

============= 

textbf{(on followup)}
how could we make these better?  regenerate the 20 decompositions so that they are higher-fidelity and are better explanations for the hypothesis.

\end{lstlisting}

\caption{Prompt used for generating fact-conditioned ad-hoc decompositions. The same prompt is re-used with an additional follow-up instruction to generate better decompositions.}
\label{fig:fact_generator_prompt}
\end{figure*}

\begin{figure*}

\paragraph{Document-conditioned entailment prompt:}  \quad \\ \vspace{-5mm}

\begin{lstlisting}[style=base]
QUESTION:
{question} 

HYPOTHESIS:
{hypothesis}

My student was trying to prove this hypothesis as it relates to the question. He pulled up this support document. 

In the context of the QUESTION, does the PASSAGE entail the HYPOTHESIS? In other words, could we reasonably infer that the HYPOTHESIS is true in the context of the QUESTION using only the information in the PASSAGE?

PASSAGE: 
{passage}

Please only make a judgment about whether the HYPOTHESIS is entailed by the PASSAGE, and not whether it answers the QUESTION. 
Please score it on a scale of 1 to 5:

1: Definitely not entailed-- PASSAGE has nothing to do with the HYPOTHESIS
2: Poorly entailed-- PASSAGE might be on topic but does not provide any evidence for the HYPOTHESIS
3: Moderately entailed-- PASSAGE provides some evidence to suggest the HYPOTHESIS is true, but there is substantial missing information or ambiguity
4: Strongly entailed-- PASSAGE provides strong evidence for the HYPOTHESIS, but there is any amount of missing information or ambiguity
5: Definitely entailed-- PASSAGE provides strong evidence for the HYPOTHESIS, and there is no missing information or ambiguity
\end{lstlisting}

\caption{Prompt used to filter and score passage-hypothesis entailment pairs.}
\label{fig:passage_entailment_prompt}
\end{figure*}

\section{Retrieval Index}
\label{app:retrieval}
We build retrieval indexes for HotpotQA and Wikipedia using the \texttt{pyserini} package \cite{Lin_etal_SIGIR2021_Pyserini}. We index all the data in HotpotQA as given in the original paper and index the 2021-01-20 version of Wikipedia with 100 word chunks. We use the BM25 algorithm \cite{robertson1995okapi} for first stage retrieval and rerank using SentenceTransformer's \texttt{ms-marco-MiniLM-L-12-v2} \cite{reimers-2019-sentence-bert}.

We retrieve the top 1000 documents and rerank and return the top 30 candidate support facts per hypothesis using as our retrieval query the concatenated question and hypothesis. 

\section{Hyperparameters}
We train the ChatGPT student for 5 epochs using the OpenAI API, and the RoBERTa student for 10 epochs using the SentenceTransformers library~\cite{reimers-gurevych-2019-sentence}. We trained separate student models on 18.3K ARC decompositions and on 20.4K Hotpot ones.

In the \sysname{} algorithm, we prompt the decomposition generators to propose 10 decompositions each, resulting in 40 candidate decompositions per hypothesis. To improve the initial search horizon, we double this number at depth 0 only. We prompt the forward-chaining prompt to produce 30 inferences entailed by retrieved documents. We use temperature=.2 sampling for all entailment filters and temperature=1 for generating decompositions. 

We set the expansion budget to 80 nodes and the filter entailment filter threshold to 0.6. We define paraphrases as having SBERT cosine similarity of 0.9 or higher.

\section{Baselines}
\begin{algorithm*}
\DontPrintSemicolon
\SetAlgoLined
\SetKwInOut{Input}{Input}
\SetKwInOut{Output}{Output}
\SetKw{Continue}{continue}
\SetKw{In}{in}

\Input{A list of queries $Q = [q_1, \ldots, q_n]$}
\Output{$\leq t$ scored trees for each query in $Q$}
\BlankLine
\ForEach{$q_i$ \In $Q$}{
    \tcp{Retrieve a set of support facts $S$ conditioned on $q_i$}
    $S \leftarrow \text{RetrieveSupportFacts}(q_i)$\;
    \tcp{Generate $m$ candidate trees using ICL Prompt}
    $T_{1}, \dots, T_t \leftarrow \text{GenerateCandidateTrees}(q_i, S, m, t)$\;
    \ForEach{$T_j$ \In $T$}{
        \tcp{Prune disconnected nodes from the tree}
        $T_j \leftarrow \text{PruneTree}(T_j)$\;
        \tcp{Check for ungrounded leaves not in $S$}
        $U \leftarrow \text{FindUngroundedLeaves}(T_j, S)$\;
        \ForEach{$l_k$ \In $U$}{
            \tcp{Check if the leaf is entailed by a fact in the full index}
            \If{\text{not IsEntailedByIndex}$(l_k)$}{
                \Continue\;
            }
        }
        \tcp{Retain tree if all leaves are grounded or entailed}
        $\text{RetainTree}(T_j)$\;
    }
}
\ForEach{\text{retained tree} $T$}{
    \tcp{grade the tree using ChatGPT (student)}
    $\text{ScoreTree}(T)$\;
}
\caption{Greedy End-to-End Entailment Tree Generator}
\label{alg:e2e_baseline}
\end{algorithm*}

    \begin{algorithm*}
\DontPrintSemicolon
\SetAlgoLined
\SetKwInOut{Input}{Input}
\SetKwInOut{Output}{Output}
\SetKw{Continue}{continue}
\SetKw{In}{in}

\Input{A list of queries $Q = [q_1, \ldots, q_n]$}
\Output{$t$ scored trees for each query in $Q$}
\BlankLine
\ForEach{$t$ \In 1, 2, \dots $t$}{
    \ForEach{$q_i$ \In $Q$}{
        \tcp{Initialize search frontier and decompositions}
        $F \leftarrow \{q_i\}$\;
        $D \leftarrow []$\;
        $N \leftarrow 0$\;
        \While{$F \neq \emptyset$ \textbf{and} $N < 10$}{
            $N \leftarrow N + 1$\;
            \tcp{Retrieve and flatten support facts for sentences in $F$}
            $S \leftarrow \text{Set}(\text{Flatten}(\text{RetrieveSupportFacts}(f) \text{ for } f \text{ in } F))$\;
            \tcp{Generate one line of tree decomposition using ICL prompt}
            $d_N \leftarrow \text{GenOneStep}(q_i, S, D)$\;
            \tcp{Append line to decompositions}
            $D \leftarrow D + [d_N]$\;
            \tcp{Create tree from decompositions}
            $T_N \leftarrow \text{CreateTree}(D)$\;
            \tcp{Prune disconnected nodes from the tree}
            $T_N \leftarrow \text{PruneTree}(T_N)$\;
            \tcp{Check for ungrounded leaves}
            $U \leftarrow \text{FindUngroundedLeaves}(T_N, S)$\;
            \ForEach{$l_i$ \In $U$}{
                \tcp{Check if the leaf is entailed by a fact $f$ in the full index}
                \If{\text{IsEntailedByIndex}($l_i, f$)}{
                    \tcp{If it is, add the entailment to the decomposition list}
                    $D \leftarrow D + [\text{``}l_i \Leftarrow f\text{''}]$\;
                    $U \leftarrow U \setminus l_i$
                }
            }
            \tcp{Set $F$ to be the remaining ungrounded leaves}
            $F \leftarrow U$\;
            
        }
    }
    Retain $T_N$
}
\ForEach{\text{retained tree} $T$}{
    \tcp{grade the tree using ChatGPT (student)}
    $\text{ScoreTree}(T)$\;
}
\caption{Stepwise Entailment Tree Generator}
\label{alg:stepwise}
\end{algorithm*}

\label{app:baselines}
\autoref{alg:e2e_baseline} depicts the pseudocode for the end-to-end tree generation baseline. \autoref{alg:stepwise} depicts the stepwise version.

\end{document}